\documentclass{article} 
\usepackage[final]{colm2025_conference}

\usepackage{microtype}
\usepackage{hyperref}
\usepackage{url}
\usepackage{booktabs}

\usepackage{lineno}
\usepackage{latexsym}

\usepackage[T1]{fontenc}
\usepackage[utf8]{inputenc}
\usepackage{graphicx}
\usepackage{xspace}
\usepackage{booktabs}
\usepackage{amsmath}
\usepackage{amssymb}
\usepackage{graphicx}
\usepackage{adjustbox}
\usepackage{multirow}
\usepackage{comment}
\usepackage{color, colortbl}
\usepackage{arydshln}
\usepackage{hhline}
\usepackage{array}
\usepackage{makecell}
\usepackage{longtable}
\usepackage{hyperref}

\usepackage{tcolorbox}
\definecolor{darkblue}{rgb}{0, 0, 0.5}
\hypersetup{colorlinks=true, citecolor=darkblue, linkcolor=darkblue, urlcolor=darkblue}

\newcommand{\rparagraph}[1]{\vspace{1.2mm}\noindent\textbf{#1.}}
\newcommand{\iparagraph}[1]{\vspace{1.2mm}\noindent\textit{#1.}}

\newcommand{\xlt}{{\textsc{xlt}}\xspace}
\newcommand{\zsxlt}{{\textsc{zs-xlt}}\xspace}

\newcommand{\ttest}{{\textsc{TTest}}\xspace}

\newcommand{\cs}{{\textsc{CS}}\xspace}
\newcommand{\mse}{{m\textsc{SE}}\xspace}

\usepackage{todonotes}

\title{Fleurs-SLU: A Massively Multilingual Benchmark for  \\ Spoken Language Understanding}

\author{Fabian David Schmidt\textsuperscript{1}, Ivan Vulić\textsuperscript{2}, Goran Glavaš\textsuperscript{1}, David Ifeoluwa Adelani\textsuperscript{3} \\
  \textsuperscript{1} Center For Artificial Intelligence and Data Science, University of Würzburg, Germany \\
  \textsuperscript{2} Language Technology Lab, University of Cambridge, United Kingdom \\
  \textsuperscript{3} Mila, McGill University and Canada CIFAR AI Chair\\
  }

\begin{document}

\ifcolmsubmission
\linenumbers
\fi

\maketitle

\begin{abstract}
    Spoken language understanding (SLU) is indispensable for half of all living languages that lack a formal writing system. Unlike for high-resource languages, for these languages, we cannot offload semantic understanding of speech to the cascade of automatic speech recognition (ASR) and text-based large language models (LLMs). Even if low-resource languages possess a writing system, ASR for these languages remains unreliable due to limited bimodal speech and text training data.  
Nonetheless, the evaluation of multilingual SLU is limited to shallow tasks such as intent classification or language identification. This is why we present Fleurs-SLU, a multilingual SLU benchmark that encompasses (i) 692 hours of speech for topical utterance classification in 102 languages and (ii) multiple-choice question answering via listening comprehension spanning 944 hours of speech across 92 languages. We extensively evaluate end-to-end speech classification models, cascaded systems that combine speech-to-text transcription with subsequent LLM-based classification, and multimodal speech-LLMs on Fleurs-SLU. Our results show that cascaded systems are more robust in multilingual SLU, though well-pretrained speech encoders can perform competitively in topical speech classification. Closed-source speech-LLMs match or surpass the performance of cascaded systems. We observe a strong correlation between robust multilingual ASR, effective speech-to-text translation, and strong multilingual SLU, indicating mutual benefits between acoustic and semantic speech representations.
\footnote{The datasets \& per-language results are available on Huggingface: \href{https://huggingface.co/datasets/WueNLP/sib-fleurs}{SIB-Fleurs}; \href{https://huggingface.co/datasets/WueNLP/belebele-fleurs}{Belebele-Fleurs}.} 

\end{abstract}

\section{Introduction}
\label{sec:introduction}
Only about half of the world’s living languages possess a formal writing system, underscoring the need for massively multilingual speech technology \citep{ethnologue}. Spoken language understanding (SLU) is hence a critical feature of inclusive technology for these languages, which cannot rely on combining automatic speech recognition (ASR) with language models to offload semantic speech understanding. However, the datasets currently available for evaluating SLU in truly relevant languages suffer from significant limitations. For instance, the Minds14 benchmark assesses SLU on 14 exclusively high-resource languages for intent classification, a task that often requires only shallow semantic processing, such as detecting specific keywords \citep{gerz2021minds14}. Likewise, while SpeechTaxi focuses on spoken topical classification of Bible verses in 28 diverse languages, SLU of Bible verse arguably does not reflect real-world usage \citep{keller2024speechtaxi}. 

Moreover, even when a low-resource language has a writing system, multilingual ASR often struggles to reliably transcribe that language due to the limited availability of bimodal speech and text data. Pre-trained on the self-supervised wav2vec-BERT objective~\citep{wav2vec2}, these models embed acoustic features that capture phonetic rather than semantic information \citep{choi2024selfsupervisedspeechrepresentationsphonetic}. Conversely, multilingual speech models that excel in semantic encoding tend to be more robust in ASR tasks by, e.g., leveraging context-related cues or cross-lingual similarities such as cognates or similar syntactic structures (cf. \S\ref{subsec:further-analyses-and-discussion}). This underscores that robust SLU is instrumental for genuinely inclusive speech technology.

To address the shortcomings of multilingual SLU evaluation of prior benchmarks, we compile Fleurs-SLU by realigning datasets derived from Flores ~\citep{nllbteam2022language}, a machine-translation benchmark with parallel sentences for over 200 languages. We first filter out silent instances from Fleurs \citep{fleurs2022arxiv} and then map the remaining data back to Flores.
The resulting dataset is then merged with SIB-200 \citep{adelani-etal-2024-sib} and Belebele \citep{bandarkar2023belebele}, that both are based on Flores, to create a topical utterance classification benchmark for 102 languages, and dataset for multiple-choice question answering (QA) on spoken paragraphs for 92 languages, respectively.

\rparagraph{Contributions} 
\textbf{1)} To the best of our knowledge, we are the first to present a multilingual SLU benchmark that spans over 100 languages and enables end-to-end SLU evaluation of multilingual speech encoders (\mse). SIB-Fleurs supports utterance classification for 7 topics with 692 hours of speech spanning 102 languages, over 70 more additional languages than existing benchmarks. Belebele-Fleurs provides textual multiple-choice QA from spoken paragraphs, totaling 944 hours of speech across 92 languages and exceeding concurrent work by 18 languages \citep{2m-belebele}.
\textbf{2)} We extensively benchmark state-of-the-art speech models on SIB-Fleurs and Belebele-Fleurs. Like prior work, we observe that Cascaded Systems (\cs) remain more robust than \mse, enabling SLU competitive with text-based language understanding \citep{keller2024speechtaxi}. We additionally benchmark state-of-the-art speech-LLMs on multilingual SLU and find that often perform on par or better than \cs. We are the first to show that \mse pre-trained on language understanding objectives can yield performance competitive to that of \cs on speech classification. 
We further isolate utterance quality as an important factor in multilingual SLU and show that, in zero-shot cross-lingual transfer (\zsxlt), worse utterance quality can substantially deteriorate transfer performance.
\textbf{3)} We empirically demonstrate that strong multilingual SLU coincides with much more robust multilingual ASR and higher-quality speech-to-English-text translation (S2ETT). This suggests that the pre-training of multilingual speech models may take SLU into account to make multilingual ASR more robust.

\section{Related Work}
\label{sec:related-work}
\paragraph{Multilingual Speech Representation Learning.}
Modern speech representation models are pre-trained with self-supervised objectives that are optionally followed-up by ASR and speech-to-text translation (S2TT) training. Among these models, mHubert employs the self-supervised wav2vec 2.0 objective on 90K hours of speech across 147 languages by predicting pseudo-labels derived from clustering raw speech features \citep{boito2024mhubert}. MMS-1b is also pretrained with wav2vec 2.0 on a large corpus of 491K hours of speech spanning 1,406 languages \citep{mms}. 
 Whisper-v3 is a Transformer encoder-decoder that has been multi-task pre-trained on multilingual ASR, speech-to-English-text-translation (S2ETT), spoken language identification, and Voice activity detection on 680k hours of audio \citep{radford2022whisper}. 
 SeamlessM4Tv2-Large is a multilingual multimodal translation model \citep{seamless2023}. It combines a text-to-text translation (T2TT) model pre-trained on 105 languages and a Conformer speech encoder pre-trained with w2v-BERT 2.0 on 4.5M hours of audio. The model is then trained on T2TT, S2TT, ASR, and knowledge distillation (KD) objectives. The authors perform KD from the text encoder to the speech encoder by minimizing the KL-divergence between the decoder's token output distributions on bi-modal speech and text data.

\rparagraph{Multilingual SLU} The evaluation of multilingual SLU is constrained to a limited set of tasks. SLU has been significantly shaped by task-oriented dialogue (ToD), with datasets created focusing on ToD-specific tasks such as intent classification and slot filling. These tasks frequently require only basic semantic understanding, often reducing to merely detecting specific keywords. Additionally, commonly used utterance-level SLU tasks like language identification (LID) and sentiment classification do not assess content-based understanding but instead rely on phonetic or prosodic features of speech encodings. As a result, the majority of SLU datasets are predominantly in English. Multilingual exceptions are otherwise limited. The Minds14 dataset for intent classification only includes 14 high-resource languages \citep{gerz2021minds14}. SpeechTaxi offers spoken topical classification for Bible verses in 28 diverse languages, which however do not adequately represent real-world domains \citep{keller2024speechtaxi}.

\rparagraph{Concurrent Work} \citet{2m-belebele} concurrently released the multilingual SLU benchmark \textsc{2M-Belebele} for 74 languages. In this dataset, the authors first extend Fleurs by approx. 20\% by incorporating human recordings for sentences that are part of Flores but were missing in Fleurs, as well as the questions and answers from Belebele. \citet{2m-belebele} find that a \cs-based approach on transcribing speech with Whisper-v3-Large and subsequently prompting Llama 3 70B with the transcription trails prompting the LLM on clean text by approx. 8 percentage points (pp).

\section{Fleurs-SLU}
\label{sec:dataset-creation}
We create Fleurs-SLU, a massively multilingual SLU benchmark for speech classification and multiple-choice QA from spoken paragraphs, from datasets that are based on Flores. 

\subsection{Core Datasets}
\label{subsec:datasets}

\textbf{Flores} consists of professionally translated 3,001 sentences of English Wikipedia paragraphs to evaluate machine translation \citep{nllbteam2022language}.\footnote{Flores comprises 3,001 sentences divided into \textsc{dev} (997 sentences), \textsc{devtest} (1,012 sentences), and \textsc{test} (992 sentences) sets. The authors did not release the \textsc{test} set.}
\textbf{Fleurs} comprises 2.3 spoken utterances, on average, per sentence from the \textsc{dev} and \textsc{devtest} splits of 102 languages in Flores \citep{fleurs2022arxiv}.\footnote{Almost all languages part of Fleurs however are missing a few hundred sentences from Flores.} Fleurs is used to evaluate multilingual ASR, LID, as well as speech-to-text and text-to-speech tanslation in all language directions.
\textbf{SIB-200} refined the topical metadata annotations of sentences in the \textsc{dev} and \textsc{devtest} splits of Flores into 7 categories \citep{adelani-etal-2024-sib}.\footnote{"science/technology", "travel", "politics", "sports", "health", "entertainment", "geography".} The resulting SIB-200 is a topical classification benchmark for 205 language variants.
\textbf{Belebele} is a multiple-choice reading comprehension benchmark for 122 languages~\citep{bandarkar-etal-2024-belebele}. The authors reconstruct paragraphs from Flores sentences. They generate 1-2 questions per English paragraph, which are professionally translated into 121 languages. Belebele comprises 900 questions that span across 488 passages.

\subsection{Benchmark Creation}
\label{subsec:benchmark-creation}

We compile Fleurs-SLU by carefully aligning data from the above benchmarks as follows. 

\rparagraph{1) Merging Fleurs \& Flores} We begin by removing silent instances from Fleurs.\footnote{Counts of removed examples are listed in Appendix \ref{subsec:silent-fleurs-examples}.} We first normalize the loudness to a target RMS level of ${-}25$ dB. We next apply voice activity detection using Silero-VAD \citep{silero}.\footnote{\url{https://github.com/snakers4/silero-vad}} Samples are deemed silent if speech is detected in less than $5\%$ of their duration.\footnote{The Silero-VAD pipeline also frequently removes inaudibly noisy samples as a side effect.}  Lastly, we verify our approach on 50 randomly sampled predicted silent instances. We find only one borderline misclassified example, which is noisy but comprehensible. We then conservatively merge Fleurs and Flores by matching instances first on exact string match and then by Levenshtein distance of 3 on normalized strings.\footnote{Normalized strings remove characters that are Unicode punctuation codepoints.}

\rparagraph{2a) SIB-Fleurs} For each language, we pool instances from the training, validation, and test splits of SIB-200~\citep{adelani-etal-2024-sib} and align the data with our merged Fleurs-Flores dataset with the same string alignment procedure as before. The data is then regrouped into the training, validation, and test splits of the original Fleurs dataset. This segmentation ensures compatibility for speech models that may be trained on ASR using the training set of Fleurs prior to SIB-Fleurs evaluation. This also ensures that the speakers in the training set are different from those in the validation and test sets. Table \ref{tab:sib-stats} lists aggregated statistics on the instance- and the utterance-level for SIB-Fleurs. Appendix \ref{subsec:sib-samples} provides a list of samples by split per language.

\begin{table}[h]
    \centering
    \small
    \begin{minipage}{0.48\linewidth}
        \centering
        \begin{tabular}{l c}
            \toprule
            \textbf{Utterance} & \\
            \textbf{Classification} & \\ \midrule
            Languages & 102 \\
            Classes (Topics) & 7 \\ \midrule
            Utterances per sample & 2.2 (2.0) \\ 
            Duration per utterance (s) & 12.9 (11.6) \\ 
            Total audio (hr) &  692 \\ 
            \hline
            \multicolumn{2}{l}{\textit{Samples by Split}} \\
            Training & 696 (728) \\
            Validation & 66 (70) \\
            Test  & 163 (174) \\
            \bottomrule
        \end{tabular}
        \caption{\textbf{Statistics of SIB-Fleurs.} Utterance-level metrics are aggregated by language and then pooled over languages. \textbf{Metrics:} either sums or averages (median).}
        \label{tab:sib-stats}
    \end{minipage}
    \hfill
    \begin{minipage}{0.48\linewidth}
        \centering
        \begin{tabular}{l c}
            \toprule
            \ \textbf{Multiple-Choice QA} & \\
            \textbf{Spoken Paragraphs} & \\ \midrule
            Languages               & 92 \\
            Answer Choices                 & 4 \\ \midrule
            Questions per paragraph        & 1.8 (1.8) \\ 
            Sentences per paragraph        & 3.6 (4.0) \\ 
            Utterances per sentence        & 2.0 (2.0) \\
            Duration per utterance (s)     & 12.6 (12.0) \\ 
            Duration per paragraph (s)     & 47.0 (43.4) \\
            Total audio (hr) & 944 \\
            \hline
            Samples (Paragraphs)  & 709 (771) \\
            \bottomrule
        \end{tabular}
        \caption{\textbf{Statistics of Belebele-Fleurs.} Metrics are aggregated by language and then pooled over languages. \textbf{Metrics:} either sums or averages (median).}
        \label{tab:belebele-stats}
    \end{minipage}
    \vspace{-0.3cm}
\end{table}

\rparagraph{2b) Belebele-Fleurs} We merge our Fleurs-Flores sentences with Belebele paragraphs by intersecting the URLs of the texts. We discard all paragraphs that are not complete in Fleurs-Flores. We verify our reconstructed paragraphs against the original Belebele by ensuring that the Levenshtein distance for strings with removed punctuation is negligibly small (less than 3 characters). Table \ref{tab:belebele-stats} provides a summary statistics on both the paragraph- and the sentence-level for Belebele-Fleurs, while Appendix \ref{subsec:belebele-fleurs} lists the samples per language. 

\section{Experimental Setup}
\label{sec:experimental-setup}
\subsection{Tasks and Languages}
\label{subsec:tasks-and-languages}

\rparagraph{SIB-Fleurs} We train \mse on the utterances and \cs on the transcriptions of the English training set. For both \mse and \cs, we feed the sequence-level representation pooled from token or speech frame embeddings into a classification head. Roberta-Large uses the \texttt{[CLS]} token as a sequence-level embedding. All other models average the token or speech frame output embeddings.

\rparagraph{Belebele-Fleurs}
We train and validate \cs models on the English training and dev sets of Belebele \citep{bandarkar-etal-2024-belebele}, respectively. We jointly embed the paragraph, question, and choices with text encoders. We then average the token encodings of each choice $c_i \in C$ and project the choice embedding via head $H^{D \times 1}$ to a logit $\mathbf{l_{c_i}}$. We minimize the cross-entropy of the concatenated choice logits $\{\mathbf{l_{c_i}}\}_{i=1}^{|C|}$ to the label choice.\footnote{We do not evaluate speech LLMs as existing models are based on Whisper-v3 which is limited to 30 seconds of audio input. We leave such evaluation to future work.}

 \subsection{Cross-Lingual Transfer Setups}

We experiment on two commonly used cross-lingual transfer (\xlt) paradigms. Zero-shot cross-lingual transfer (\zsxlt) and Translate-Test (\ttest) allow us to evaluate \xlt\ without requiring additional annotation for any target language. In \zsxlt, we first train a multilingual model on the English source-language data (cf. \S\ref{subsec:tasks-and-languages}) and then directly run inference on the target-language test instances. In \ttest, the model is also first fine-tuned on labeled English source-language data. At test time, the target-language examples are translated to the source language prior to inference, which enables \xlt with monolingual LLMs. We additionally evaluate zero-shot prompting of speech-LLMs. Here, we provide the context of the task in English and all relevant input in-language in the respective modality, i.e. text or utterances, to the speech-LLMs (cf. \S\ref{appendix:speech-llms}). 

\rparagraph{Speech Classification} We evaluate state-of-the-art \mse for speech classification (cf. \S\ref{sec:related-work}). 
 We include MMS-1B without fine-tuning (`MMS-1B'), with ASR fine-tuning on Fleurs (`MMS-1B-Fleurs'), and with ASR fine-tuning on multilingual datasets (`MMS-1B-all'), allowing us to analyze the impact of ASR fine-tuning on cross-lingual SLU. For ASR fine-tuning, \citet{mms} train language adapters and language-specific decoding heads while keeping all other parameters frozen. For task fine-tuning, we freeze the adapters to facilitate \zsxlt. We further evaluate mHubert and the speech encoders of both Whisper-v3-Large and SeamlessM4Tv2-Large (cf. \S\ref{sec:related-work}).

\rparagraph{Cascading \& Text}
Cascaded systems (\cs) perform \xlt\ in two steps: (i) an ASR model transcribes speech into text, and (ii) a text encoder processes the transcription via a classification head. We consider two transcription targets: the target language (in-language) and English (speech-to-English translation, S2ETT). S2ETT corresponds to \ttest, enabling \xlt\ with monolingual LLMs.
We use SeamlessM4Tv2-Large and Whisper-v3-Large as ASR backends. For languages not supported by the model, we manually select the closest available language for in-language transcription. All decoding is greedy, as more complex decoding strategies do not improve \xlt\ performance \citep{ebing-glavas-2024-translate}.
We additionally evaluate on the original text (\textsc{Text}) of SIB-200 and Belebele. For \ttest, we translate this text to English using SeamlessM4Tv2-Large.\footnote{We use the gold Q\&A in English for \ttest. Otherwise, MT models would need to be combined, e.g., S2ETT of paragraphs with Whisper, and translation of textual Q\&A with SeamlessM4Tv2.}
For classification, we evaluate three text encoders: Roberta-Large \citep{liu2019roberta}, LLM2Vec \citep{llm2vec}, and NLLB-LLM2Vec \citep{schmidt-etal-2024-self}. LLM2Vec is a sequence encoder based on Llama 3 8B \citep{llama3}, trained with bidirectional attention on masked next-token prediction and SimCSE \citep{gao-etal-2021-simcse}. NLLB-LLM2Vec extends LLM2Vec with the encoder of the NLLB translation model, covering 200+ languages for robust multilingual NLU \citep{schmidt-etal-2024-self}.

\rparagraph{Speech-LLMs}
We further evaluate two multimodal speech-LLMs, Qwen 2.5 7B-Omni \citep{QwenOmni} and Gemini 2.0 Flash\footnote{\url{https://cloud.google.com/vertex-ai/generative-ai/docs/models/gemini/2-0-flash}}, in a zero-shot prompting setup.\footnote{Note that the audio encoder of Qwen 2.5 7B-Omni is initialized with Whisper-v3-Large prior to further cross-modal instruction tuning.} For each task, we provide an English task description and supply the task-specific input (e.g., question, paragraph, or utterance) in the target language, using the appropriate modality (text or speech).
To ensure consistent audio quality across languages, we normalize all utterances to a root mean square (RMS) level of 0.07. For Belebele-Fleurs, where inputs consist of multiple utterances, we normalize the concatenated sequence. Appendix~\S\ref{appendix:speech-llms} details the prompt formats and input preprocessing used for each task and modality.

\subsection{Further Details}

\rparagraph{Hyperparameters}
We train all \mse and \cs models with AdamW~\citep{loshchilov2018decoupled}, weight decay of $0.01$, and with 10\% linear warm-up and followed by linear decay on an effective batch size of 32. For SIB-Fleurs, we run a grid search over the learning rates $\{1,2,\dots,9,10\}e^{-5}$, since suitable hyperparameters have not yet been extensively studied for \mse on such a downstream task. For Belebele-Fleurs, we fine-tune the LLMs on the learning rates $\{1,2,3\}e^{-5}$ with LoRAs of rank $r{=}16$ and alpha $\alpha{=}32$ attached onto all linear layers, as full fine-tuning is prohibitively expensive. We train models for 20 and 3 epochs for SIB-Fleurs and Belebele-Fleurs, respectively, and validate at every 10\% of training steps. Experimental results are averaged across 3 random seeds. We report results for runs on the learning rate that performs best, on average, on the English validation sets. For experiments with Qwen 2.5 7B-Omni and Gemini 2-Flash, we perform greedy decoding. We report further experimental details of our prompting experiments in \S\ref{appendix:speech-llms}.

\rparagraph{Text vs. Utterance Quality} We evaluate models trained on the original text (\textsc{Text}) as well as on speech data of varying quality. Each sentence in Fleurs is associated with one or more utterances. To quantify speech quality, we compute the character error rate (CER) between the reference Flores sentence and its transcriptions produced by Whisper-v3-Large and SeamlessM4Tv2-Large.
Based on these CER scores, we construct two utterance subsets: one comprising the lowest-CER utterances (best quality), and one with the highest-CER utterances (worst quality). We fine-tune the \mse and \cs models and evaluate all models (incl. speech-LLMs) separately on each subset to isolate the effect of utterance quality on downstream performance. This setup also balances the training data to enable a fair comparison between \mse, \cs, and \textsc{Text}.

\section{Results}
\label{sec:results}

Table \ref{tab:main-results} summarizes the main results for both tasks by approach across our \xlt setups. We dissect the results along several axes of analysis.

\rparagraph{English} The first column presents the in-language English performance by task, modality (\textsc{Text} vs. \mse, \cs, and speech-LLM), and utterance quality by model.

\iparagraph{Text, CS, and Speech-LLMs} English ASR performance is strong for both transcription models (cf. Figure \ref{fig:bleu-asr-analysis}), and \cs effectively leverages the NLU capabilities of LLMs, thanks to their extensive pre-training on both tasks. \cs are even on par with and sometimes outperform models trained on gold text, underscoring the quality of ASR on English. The slight outperformance of \cs over \textsc{Text} in SIB-Fleurs results from model selection on comparatively small validation splits (cf. Table \ref{tab:sib-stats}). As a result, \cs outperforms all \mse models on SIB-Fleurs. On Belebele-Fleurs, both \cs models backed by LLM2Vec and NLLB-LLM2Vec perform comparably. In sum, the transcription model has no clear impact on English performance. The speech-LLMs perform on par or better on Belebele-Fleurs while slightly trailing the best fine-tuned models on both \cs and \textsc{Text} on SIB-Fleurs. Gemini 2.0 Flash consistently performs better than Qwen 2.5B 7B-Omni. Notably, their cross-modal gap (\textsc{Text} vs. speech-LLM) is small (approx. -1.5\%).

\begin{table*}[t!]
\begin{adjustbox}{width=\textwidth,center}
    \renewcommand{\arraystretch}{0.9} 
    \begin{tabular}{@{\extracolsep{\fill}} 
        c!{\vrule width \arrayrulewidth} 
        l!{\vrule width \arrayrulewidth} 
        c!{\vrule width \arrayrulewidth} 
        c!{\vrule width \arrayrulewidth} 
        >{\columncolor{gray!25}}c!{\vrule width \arrayrulewidth} 
        c!{\vrule width \arrayrulewidth} 
        >{\columncolor{gray!25}}c!{\vrule width \arrayrulewidth} 
        c!{\vrule width \arrayrulewidth} 
        >{\columncolor{gray!25}}c!{\vrule width \arrayrulewidth} 
        c!{\vrule width \arrayrulewidth} 
        >{\columncolor{gray!25}}c!{\vrule width \arrayrulewidth} 
        c!{\vrule width \arrayrulewidth} 
        >{\columncolor{gray!25}}c}
        \toprule
        \multicolumn{3}{l}{ } &\multicolumn{2}{c}{\textbf{English}} & \multicolumn{8}{c}{\textbf{Non-English}} \\
        \cmidrule(lr){4-5} \cmidrule(lr){6-13}

        \multicolumn{3}{l!{\vrule width \arrayrulewidth}}{\textbf{\textit{Utterance Quality}} (\textbf{B}est; \textbf{W}orst)}   &  
        \textbf{B} & \textbf{W} & \textbf{B} & \textbf{W} & \textbf{B} & \textbf{W} & \textbf{B} & \textbf{W} & \textbf{B} & \textbf{W} \\
        \midrule 
        \multicolumn{2}{l!{\vrule width \arrayrulewidth}}{\textbf{\textit{Language Groups (Size)}}} & \textbf{Setup} & 
         \multicolumn{2}{c!{\vrule width \arrayrulewidth}}{\textbf{EN}} & \multicolumn{2}{c!{\vrule width \arrayrulewidth}}{\textbf{Whisper}} & \multicolumn{2}{c!{\vrule width \arrayrulewidth}}{\textbf{S4T}} & \multicolumn{2}{c!{\vrule width \arrayrulewidth}}{\textbf{Unsup.}} & \multicolumn{2}{c}{\textbf{Non-\textsc{EN}}} \\
        \hline
        \rowcolor{green!10} \multicolumn{3}{l!{\vrule width \arrayrulewidth}}{\rule{0pt}{2.5ex} \textbf{\textit{SIB-Fleurs}}} & \multicolumn{2}{c!{\vrule width \arrayrulewidth}}{\textbf{(1)}} & \multicolumn{2}{c!{\vrule width \arrayrulewidth}}{\textbf{(85)}} & \multicolumn{2}{c!{\vrule width \arrayrulewidth}}{\textbf{(90)}} & \multicolumn{2}{c!{\vrule width \arrayrulewidth}}{\textbf{(7)}} & \multicolumn{2}{c}{\textbf{(101)}} \\
        \hline \rule{0pt}{2.5ex} 
        \multirow{5}{*}{\shortstack{\textbf{mSE}}} 
        & \textsc{mHuBERT}              & X & 41.1 & 39.7 & 27.3 & 19.1 & 27.2 & 19.1 & 26.5 & 16.3 & 26.9 & 18.6 \\
        & \textsc{MMS-1b}               & X & 44.6 & 46.9 & 19.9 & 13.6 & 19.8 & 13.5 & 20.8 & 13.3 & 19.8 & 13.4 \\
        & \textsc{MMS-1b-Fleurs}        & X & 64.8 & 64.2 & 25.2 & 20.8 & 24.7 & 20.3 & 27.1 & 22.1 & 25.0 & 20.6 \\
        & \textsc{MMS-1b-All}           & X & 55.2 & 53.1 & 21.7 & 18.1 & 21.4 & 17.9 & 24.9 & 21.0 & 21.7 & 18.2 \\ 
        & \textsc{Whisper-v3-L}         & X & 78.3 & 76.3 & 46.0 & 42.0 & 45.5 & 41.3 & 39.4 & 38.4 & 44.7 & 40.8 \\
        & \textsc{SeamlessM4Tv2-L}      & X & 87.4 & 88.5 & \underline{82.3} & 79.3 & 82.2 & 79.1 & 55.7 & 52.9 & 79.0 & 75.8 \\
 
        \hline \rule{0pt}{2.5ex} 
        \multirow{6}{*}{\shortstack{\textbf{CS}}} & 
          \textsc{Roberta$_{\text{Large}}$-Wh-EN}           & T & 90.4 & 91.7 & 75.8 & 72.7 & 73.6 & 70.8 & 52.0 & 48.7 & 71.5 & 68.4 \\ 
        & \textsc{Roberta$_{\text{Large}}$-S4T-EN}          & T & 91.7 & 90.8 & 86.2 & 84.5 & 85.9 & \underline{84.2} & 56.3 & 55.2 & \underline{82.3} & 80.6 \\
        & \textsc{LLM2Vec-Wh-EN}             & T & 91.1 & \underline{92.3} & 76.6 & 75.3 & 74.5 & 73.4 & 53.0 & 53.2 & 72.4 & 71.1 \\
        & \textsc{LLM2Vec-S4T-EN}            & T & \underline{93.0} & 91.3 & \textbf{87.1} & \textbf{85.2} & \textbf{86.9} & \textbf{84.9} & 58.1 & \underline{56.6} & \textbf{83.4} & \underline{81.4} \\
        & \textsc{NLLB-LLM2Vec-Wh}           & X & 92.5 & 91.0 & 77.5 & 73.9 & 75.4 & 72.2 & 54.4 & 53.2 & 73.7 & 70.4 \\
        & \textsc{NLLB-LLM2Vec-S4T}          & P & \textbf{94.0} & \textbf{92.7} & \underline{85.1} & \underline{84.9} & \underline{85.3} & \textbf{84.9} & \textbf{62.3} & \textbf{60.7} & \underline{82.3} & \textbf{81.9} \\ \hline \rule{0pt}{2.5ex}
        \textbf{Speech-} & \textsc{Qwen 2.5 7B-Omni}          & P & 87.0 & 87.6 & 59.5 & 58.4 & 58.3 & 57.3 & 45.6 & 45.6 & 56.9 & 55.9 \\  
        \textbf{LLM} & \textsc{Gemini 2.0 Flash}         & P & 87.6 & 88.7 & 82.1 & 81.4 & 80.8 & 80.0 & \underline{59.4} & 59.3 & 78.6 & 77.8 \\ \hline \rule{0pt} {2.5ex}
        \multirow{5}{*}{\shortstack{\textsc{\textbf{Text}}}} & 
          \textsc{Roberta$_{\text{Large}}$-S4T-EN}  & T & \multicolumn{2}{c!{\vrule width \arrayrulewidth}}{\textbf{92.3}}   & \multicolumn{2}{c!{\vrule width \arrayrulewidth}}{87.4} & \multicolumn{2}{c!{\vrule width \arrayrulewidth}}{\textbf{89.1}} & \multicolumn{2}{c!{\vrule width \arrayrulewidth}}{55.3} & \multicolumn{2}{c}{85.2} \\ 
        & \textsc{LLM2Vec-S4T-EN}   & T & \multicolumn{2}{c!{\vrule width \arrayrulewidth}}{\underline{91.0}} & \multicolumn{2}{c!{\vrule width \arrayrulewidth}}{86.4} & \multicolumn{2}{c!{\vrule width \arrayrulewidth}}{\underline{87.9}} & \multicolumn{2}{c!{\vrule width \arrayrulewidth}}{55.6} & \multicolumn{2}{c}{84.1} \\
        & \textsc{NLLB-LLM2Vec} & X & \multicolumn{2}{c!{\vrule width \arrayrulewidth}}{\textbf{92.3}} & \multicolumn{2}{c!{\vrule width \arrayrulewidth}}{\textbf{88.1}} & \multicolumn{2}{c!{\vrule width \arrayrulewidth}}{\underline{87.9}} & \multicolumn{2}{c!{\vrule width \arrayrulewidth}}{\textbf{80.7}} & \multicolumn{2}{c}{\textbf{87.2}} \\
        
        & \textsc{Qwen 2.5 7B-Omni} & P & \multicolumn{2}{c!{\vrule width \arrayrulewidth}}{88.7} & \multicolumn{2}{c!{\vrule width \arrayrulewidth}}{76.4} & \multicolumn{2}{c!{\vrule width \arrayrulewidth}}{75.1} & \multicolumn{2}{c!{\vrule width \arrayrulewidth}}{53.9} & \multicolumn{2}{c}{73.0} \\
        & \textsc{Gemini 2.0 Flash} & P & \multicolumn{2}{c!{\vrule width \arrayrulewidth}}{90.4} & \multicolumn{2}{c!{\vrule width \arrayrulewidth}}{\underline{87.0}} & \multicolumn{2}{c!{\vrule width \arrayrulewidth}}{86.8} & \multicolumn{2}{c!{\vrule width \arrayrulewidth}}{\underline{77.2}} & \multicolumn{2}{c}{\underline{86.1}} 
        \\ 
        
        \midrule \hline 
          \rowcolor{green!10} \multicolumn{3}{l!{\vrule width \arrayrulewidth}}{\rule{0pt}{2.5ex} \textbf{\textit{Belebele-Fleurs}}} & \multicolumn{2}{c!{\vrule width \arrayrulewidth}}{\textbf{(1)}} & \multicolumn{2}{c!{\vrule width \arrayrulewidth}}{\textbf{(79)}} & \multicolumn{2}{c!{\vrule width \arrayrulewidth}}{\textbf{(84)}} & \multicolumn{2}{c!{\vrule width \arrayrulewidth}}{\textbf{(4)}} & \multicolumn{2}{c}{\textbf{(91)}} \\
        \hline \rule{0pt}{2.5ex} 
        \multirow{6}{*}{\shortstack{\textbf{CS}}} & 
          \textsc{Roberta$_{\text{Large}}$-Wh-EN}          & T & 82.6 & 81.3 & 57.5 & 56.2 & 56.4 & 55.2 & 46.2 & 44.7 & 55.5 & 54.3 \\ 
        & \textsc{Roberta$_{\text{Large}}$-S4T-EN}         & T & 82.0 & 79.4 & 67.5 & 66.4 & 67.0 & 65.8 & 47.2 & 46.2 & 65.2 & 64.1 \\ 
        & \textsc{LLM2Vec-Wh-EN}            & T & \underline{94.9} & \textbf{94.4} & 74.6 & 73.5 & 73.6 & 72.6 & \underline{63.5} & \underline{62.9} & 72.7 & 71.8 \\ 
        & \textsc{LLM2Vec-S4T-EN}           & T & \textbf{95.5} & \textbf{94.4} & \underline{84.0} & \underline{83.1} & \textbf{83.5} & \textbf{82.5} & \textbf{64.7} & \textbf{63.6} & \textbf{81.8} & \textbf{80.9} \\ 
        & \textsc{NLLB-LLM2Vec-Wh}          & X & 94.8 & \underline{93.5} & 58.0 & 56.2 & 56.8 & 55.0 & 40.9 & 41.0 & 55.5 & 53.8 \\ 
        & \textsc{NLLB-LLM2Vec-S4T}         & X & \underline{94.9} & \underline{93.5} & 61.3 & 60.2 & 60.8 & 59.6 & 43.9 & 41.1 & 59.2 & 57.8 \\ \hline \rule{0pt}
        {2.5ex} 
        \textbf{Speech-} & \textsc{Qwen 2.5 7B-Omni}     & P & 91.6 & 91.0 & 51.4 & 50.8 & 50.4 & 49.7 & 34.4 & 34.4 & 49.0 & 48.4 \\ 
        \textbf{LLM} & \textsc{Gemini 2.0 Flash}         & P & 94.1 & 93.4 & \textbf{84.4} & \textbf{83.5} & \underline{82.9} & \underline{82.0} & 59.7 & 59.4 & \underline{81.1} & \underline{80.2} \\ \hline \rule{0pt}
        {2.5ex} 
        \multirow{5}{*}{\shortstack{\textsc{\textbf{Text}}}} & 
          \textsc{Roberta$_{\text{Large}}$-S4T-EN}           & T & \multicolumn{2}{c!{\vrule width \arrayrulewidth}}{83.3}   & \multicolumn{2}{c!{\vrule width \arrayrulewidth}}{72.1} & \multicolumn{2}{c!{\vrule width \arrayrulewidth}}{72.6} & \multicolumn{2}{c!{\vrule width \arrayrulewidth}}{48.3} & \multicolumn{2}{c}{70.5} \\ 
        & \textsc{LLM2Vec-S4T-EN}             & T & \multicolumn{2}{c!{\vrule width \arrayrulewidth}}{\underline{95.3}}   & \multicolumn{2}{c!{\vrule width \arrayrulewidth}}{\underline{87.8}} & \multicolumn{2}{c!{\vrule width \arrayrulewidth}}{\underline{88.3}} & \multicolumn{2}{c!{\vrule width \arrayrulewidth}}{\underline{66.0}} & \multicolumn{2}{c}{\underline{86.3}} \\ 
        & \textsc{NLLB-LLM2Vec}           & X & \multicolumn{2}{c!{\vrule width \arrayrulewidth}}{95.1} & \multicolumn{2}{c!{\vrule width \arrayrulewidth}}{65.6} & \multicolumn{2}{c!{\vrule width \arrayrulewidth}}{64.6} & \multicolumn{2}{c!{\vrule width \arrayrulewidth}}{55.2} & \multicolumn{2}{c}{63.8} \\ 
        & \textsc{Qwen 2.5 7B-Omni} & P & \multicolumn{2}{c!{\vrule width \arrayrulewidth}}{94.1} & \multicolumn{2}{c!{\vrule width \arrayrulewidth}}{69.1} & \multicolumn{2}{c!{\vrule width \arrayrulewidth}}{67.4} & \multicolumn{2}{c!{\vrule width \arrayrulewidth}}{36.6} & \multicolumn{2}{c}{64.9} \\
        & \textsc{Gemini 2.0 Flash} & P & \multicolumn{2}{c!{\vrule width \arrayrulewidth}}{\textbf{95.9}} & \multicolumn{2}{c!{\vrule width \arrayrulewidth}}{\textbf{90.8}} & \multicolumn{2}{c!{\vrule width \arrayrulewidth}}{\textbf{89.7}} & \multicolumn{2}{c!{\vrule width \arrayrulewidth}}{\textbf{82.0}} & \multicolumn{2}{c}{\textbf{89.1}}
        \\
        \bottomrule
    \end{tabular}
\end{adjustbox}
\vspace{-0.1cm}
\caption{\textbf{\zsxlt with \mse, \cs, and \textsc{\textbf{Text}} and zero-shot prompting with speech-LLMs (cf. \S\ref{sec:experimental-setup}).} We report accuracy averaged over 3 seeds on checkpoints that maximize perf. on English validation sets. \cs: The suffixes \textsc{Wh} and \textsc{S4T} denote Transcription \& Translation (incl. \textsc{-En}: \{Text,Speech\}-to-English-Text-Translation), respectively. Numbers in parentheses denominate size of group (e.g., Whisper-v3 supports 84 languages of SIB-Fleurs). \textbf{Setup}: X=Zero-shot Cross-Lingual Transfer;T=Translate-Test;P=Zero-Shot Prompting (cf. \S\ref{sec:experimental-setup}). \textbf{Abbreviations}: EN=English;Whisper=Whisper-v3-Large;S4T=SeamlessM4Tv2-Large; Unsup.=Unsupported. The (second-)best model in each column in \textbf{bold} (\underline{underline}).}
\label{tab:main-results}
\vspace{-0.6cm}
\end{table*}

\iparagraph{mSE} The English results on SIB-Fleurs suggest that the SLU capabilities of \mse models are strongly shaped by their pre-training curriculum. mHubert (41.1\%) and MMS-1B (44.6\%), which are pre-trained solely in a self-supervised manner, underperform other \mse on SIB-Fleurs. Fine-tuning MMS-1B on ASR, either on Fleurs (MMS-1B-Fleurs) or Fleurs combined with additional data (MMS-1B-All), results in significant improvements on SIB-Fleurs of +10-20 percentage points (pp). Notably, MMS-1B-Fleurs outperforms MMS-1B-All (+9.6\,pp), suggesting that the broader domain mixture in the MMS-1B-All training set negatively affects this task. Whisper-v3-Large surpasses all MMS-1B variants (78.3\%). The large-scale pre-training of Whisper-v3 on both multilingual ASR and S2ETT enhances its SLU performance. SeamlessM4Tv2-Large is the best performing \mse on the English test set of SIB-Fleurs (87.4\%), with only a slight performance gap compared to \cs (approx. -3\,pp). The joint pre-training on multilingual ASR as well as multilingual and cross-modal MT with text-to-speech knowledge distillation (KD) significantly enriches the semantics in speech representations of SeamlessM4Tv2-Large. The results particularly highlight that (i) multilingual ASR, (ii) S2ETT pre-training, and (iii) text-to-speech KD are crucial training objectives for enabling speech encoders to acquire strong SLU capabilities.

\iparagraph{Utterance Quality} The quality of English utterances does not affect performance of the models across both tasks. This can be attributed to the large-scale training on English ASR data of all models. This is why \cs are also on par with \textsc{Text}.

\rparagraph{\zsxlt} The right-hand side of Table \ref{tab:main-results} reports \zsxlt performance. For each task, we group the languages into (i) languages supported by Whisper-v3, (ii) languages supported by SeamlessM4Tv2, and (iii) languages unsupported by either model. Whisper-v3 and SeamlessM4Tv2 overlap in their support for 81 Fleurs languages.\footnote{For unsupported languages, we hand-select the closest supported language to transcribe into.}

\iparagraph{Text, CS, and Speech-LLMs} 
Across both tasks and all language groups, models trained on transcriptions of SeamlessM4Tv2-Large consistently outperform models fine-tuned on transcriptions of Whisper-v3-Large. The performance gap grows as Whisper's support decreases or becomes unavailable for the target languages (cf. `S4T' and `Unsup.' language groups in Table \ref{tab:main-results}). Nevertheless, while \cs are mostly competitive on SIB-Fleurs, they trail models evaluated on ground-truth paragraphs more significantly on Belebele-Fleurs (approx. -5\,pp). We presume that transcription and S2ETT errors propagate more severely in multiple-choice QA. Moreover, the best speech-LLM, Gemini 2.0 Flash, performs highly competitively across tasks, boasting on par or better performance than other models.

For SIB-Fleurs, all \cs with SeamlessM4Tv2-Large deteriorate only slightly in \xlt performance to all 101 target language, on average, relative to English (approx. -10\,pp). 
Most of this \xlt gap comes from the 7 target languages that SeamlessM4Tv2 does not support (approx. -35\,pp on average). In contrast, pairing LMs with Whisper-v3-Large causes more pronounced drops on languages that Whisper supports (approx. -13\,pp) and again a larger deficit on languages neither model supports (approx. -32\,pp). \cs are highly competitive to \textsc{Text} (approx. -0.7\,pp). Only NLLB-LLM2Vec evaluated on gold text tremendously outperforms all other models on unsupported languages, since they are supported by NLLB \citep{nllbteam2022language}.

For Belebele-Fleurs, S2ETT paired with LLM2Vec (i.e., \ttest) outperforms \zsxlt cascading on in-language transcriptions and NLLB-LLM2Vec, except for unsupported languages. This suggests that S2ETT of both Whisper-v3 and SeamlessM4Tv2 sufficiently translates the target languages into English for successful NLU. On unsupported languages, however, NLLB-LLM2Vec performs better for likely two reasons. First, for languages with low S2ETT quality, in-language transcriptions to closely related languages likely better preserve the core meanings of input sequences. Second, in addition to translation, NLLB was pre-trained with denoising autoencoding, making NLLB-LLM2Vec more resistant to noisy inputs than LLM2Vec. Overall, the findings suggest that SeamlessM4Tv2-Large is more robust for both in-language ASR as well as S2ETT (cf. Figure \ref{fig:bleu-asr-analysis}).


The performance of speech-LLMs varies substantially across models and language groups. Gemini 2.0 Flash strongly outperforms Qwen 2.5 7B-Omni on both tasks in non-English settings. Qwen exhibits a marked performance degradation when transitioning from English to non-English inputs, dropping by approx. 30\,pp in \textsc{Text}, and even more steeply in the speech modality, where the cross-modal gap widens from approx. 3\% (English) to 18\% (non-English). In contrast, Gemini maintains strong performance on non-English \textsc{Text} inputs and shows only a moderate cross-modal drop of approx. 9\%. This however exceeds the gap for the best fine-tuned \cs models vs. their \textsc{Text} counterparts (approx. 3\%).\\
Qwen relies on Whisper-v3-Large as its audio encoder, making it sensitive to Whisper’s transcription limitations. Moreover, its instruction tuning possibly is focused on a small subset of high-resource languages, potentially limiting its generalization to languages not seen during alignment or fine-tuning \citep{QwenOmni}. This is consistent with its sharp drop in performance on unsupported languages. By contrast, Gemini’s closed-source nature precludes detailed analysis, but its robustness across modalities and languages suggests a more extensive and diverse instruction tuning setup. It may also benefit from large-scale multilingual pretraining, enhanced alignment objectives, or broader training coverage.

\iparagraph{mSE} The \zsxlt performance of \mse on SIB-Fleurs mirrors the trends we observed for English performance.
When trained solely with the wav2vec 2.0 objective, MMS-1B (19\%) only slightly surpasses random performance (14\%) in \zsxlt for the highest-quality utterances, regardless of the target language group. In contrast to English, post-hoc modular ASR fine-tuning of MMS-1B with language adapters and language-specific decoding heads, whether on Fleurs alone (MMS-1B-Fleurs) or on Fleurs with additional data (MMS-1B-All), yields much smaller gains in \zsxlt (approx. +1-3\,pp).
Furthermore, despite employing fully parameter-shared multilingual ASR and S2ETT pre-training, Whisper-v3-Large fails to transfer strong English in-language performance to other languages effectively, with the \xlt gap ranging from -32.3\,pp for supported languages to -38.9\,pp for unseen languages.
In contrast, only SeamlessM4Tv2-Large achieves performance comparable to English on supported languages (-5.2\,pp), though also deflates in performance on unsupported languages (55.7\%). Overall, these findings nevertheless indicate that multilingual cross-modal translation and multilingual text-to-speech distillation align and semantically enrich multilingual speech representations to enhance cross-lingual SLU.

\iparagraph{Utterance Quality} The utterance quality consistently affects performance on both tasks for all model configurations evaluated on non-English languages. Notably, \cs and speech-LLMs seem to be less affected than \mse. We attribute this to the much more sizable pre-training on text of diverse quality of LLMs compared to speech-only models. For SIB-Fleurs, the magnitude of the performance gaps between best and worst quality utterances in \zsxlt mimics how well the speech model backbones perform across both tasks in both \cs and end-to-end speech classification. SeamlessM4Tv2-Large is more robust to noisy utterances than Whisper-v3-Large, whereas MMS variants suffer from the largest drops (approx. 5\,pp).

\subsection{Further Analyses and Discussion}
\label{subsec:further-analyses-and-discussion}

\begin{figure*}[ht!]
    \includegraphics[width=\textwidth, trim={0 0 0 0.1cm},clip]{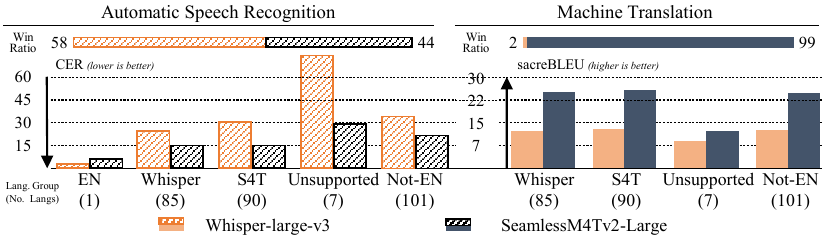}
    \vspace{-0.6cm}
    \caption{\textbf{ASR \& Speech-to-English-Text Translation (S2ETT).} CER and sacreBLEU scores for ASR and S2ETT outputs on Fleurs utterances, evaluated against original Flores sentences and pooled across all splits. Numbers in parentheses indicate the number of supported languages per model; e.g., Whisper-v3 (SeamlessM4Tv2) supports 85 (90) of the 101 non-English Fleurs languages. The two models combined do not support 7 languages. ‘Win Ratio’ denotes the number of languages on which a model outperforms the other. Abbreviations: Whisper = Whisper-v3-Large; S4T = SeamlessM4Tv2-Large.}

    \label{fig:bleu-asr-analysis}
\end{figure*}

\rparagraph{ASR \& Translation Performance} To further understand the underlying factors behind our main results (cf. Table \ref{tab:main-results}), we benchmark Whisper-v3-Large and SeamlessM4Tv2-Large on ASR and Speech-to-English-text-translation (S2ETT) on all 102 Fleurs languages in Figure \ref{fig:bleu-asr-analysis}. We first compute CER and sacreBLEU between the Flores sentences and the in-language transcriptions and S2ETT transcriptions of utterances in Fleurs pooled across all splits, respectively.\footnote{We verify that there is no leakage of dev and test splits of Fleurs in SeamlessM4Tv2-Large and Whisper-v3-Large. We observe that the averages and the standard deviations CER and sacreBLEU by task and language for either model are highly comparable across splits. We trust that the models have not been trained on the dev and test set of Fleurs.} We `macro-average' the metrics over all languages.

\vspace{1.2mm}
\noindent \textit{ASR.} Whisper-v3-Large outperforms SeamlessM4Tv2-Large for ASR on 58 out of 102 Fleurs languages (cf.~`win ratio'). Nevertheless, SeamlessM4Tv2-Large is the overall more robust transcription model: in transcription for languages \textit{other than English}, the CER of Whisper-v3-large is more than twice as high as the CER of SeamlessM4Tv2-Large, on average. 
The full per-language results (cf. Appendix \ref{subsec:asr-full-results})) confirm that SeamlessM4Tv2-Large exhibits much better ASR support for low(er)-resource languages, while being competitive in transcription quality for higher-resource languages. 

\noindent \textit{S2ETT.} SeamlessM4Tv2-Large outperforms Whisper-v3-large across the board in speech-to-English-text-translation and is favored for all but 2 languages. For both groups of languages supported by Whisper-v3-Large and SeamlessM4Tv2-Large, respectively, SeamlessM4Tv2-Large achieves on average about 8 higher sacreBLEU than Whisper-v3-Large. The gap from SeamlessM4T-Large to Whisper-v3-Large reduces to about 5 sacreBLEU for unsupported languages. This supports the notion that SeamlessM4Tv2-Large is a more robust multilingual speech encoder. 

The more robust ASR and much stronger translation performance of SeamlessM4Tv2-Large results stem from its pre-training. SeamlessM4Tv2 first initializes a text encoder and decoder with weights from a pre-trained NLLB translation model \citep{nllbteam2022language} and a speech encoder pre-trained on 4.5M hours of self-supervised training on w2v-BERT 2.0 objective. The model is then trained on translation objectives from text and speech to text between any two languages in both translation directions. The text encoder-decoder backbone is used to train the speech encoder with token-level knowledge distillation objectives on decoder output representations. On the contrary, Whisper-v3 trains models from scratch on, among others, in-language ASR and S2ETT (cf. \S\ref{sec:experimental-setup}). Consequently, the mixture of strong initialization from existing MT backbones, text-to-speech knowledge distillation, and multimodal translation objectives result in much stronger translation performance.

The main results (cf. Table \ref{tab:main-results}), together with our ASR and S2ETT analyses, underscore that sizable text and pre-training coupled with cross-modal and multilingual translation and text-to-speech knowledge distillation infuses rich semantic knowledge into \mse, as witnessed by the SLU performance of SeamlessM4Tv2-based models on Fleurs-SLU.

\rparagraph{Length Adaptor} The sequence length of encoded speech typically far exceeds the length of embedded tokenized text for the same input. The \mse of SeamlessM4Tv2-Large thus appends a temporal convolution, a `length adaptor', as its final layer to reduce the resolution of speech frames by a factor of 8 and to better align the modalities \citep{seamless2023}.\footnote{For architectural details, we refer to the original paper \citep{seamlessv1}.} 
This begs the question whether a length adaptor is essential for pooling semantics from speech output tokens. We hence compare SeamlessM4Tv2-Large's performance on SIB-Fleurs with and without the length adaptor.

Table \ref{tab:s4t-length-adaptor} presents the inconclusive results for both variants. For English, performance improves on high-quality utterances but decreases on lower-quality ones. In contrast, performance most pronouncedly declines for low-quality utterances in \zsxlt to supported languages (-2.6\%). The \zsxlt unsupported languages is not affected for high-quality utterances (-0.1\%) and only slightly for low-quality ones (-0.9\%). Three factors may explain this modest gap. First, removing the length adaptor reduces model size by 47M parameters (-7.4\,pp), which were part of substantial pre-training. Second, the length adaptor, as the final layer of the \mse, is most explicitly trained to embed speech into a shared multilingual space, attended to by the text decoder at every layer. Finally, the length adaptor might filter noisy frames through temporal downsampling to improve the robustness of speech encodings. While a modest gap persists across setups, we cannot decisively infer whether appending a length adaptor is, as opposed to replacing model capacity with other layers of equal parameter size during pre-training, crucial for improved multilingual SLU. We conclude that the pre-training regime is more vital for multilingual SLU capabilities of \mse than nuanced architectural design choices. We leave a more detailed investigation into the utility of length adaptors in SLU to future work.

\begin{table}[t]
\small
    \begin{center}
    \renewcommand{\arraystretch}{0.9} 
    \begin{tabular}{@{\extracolsep{\fill}} 
        l!{\vrule width \arrayrulewidth} 
        c!{\vrule width \arrayrulewidth} 
        c!{\vrule width \arrayrulewidth} 
        >{\columncolor{gray!25}}c!{\vrule width \arrayrulewidth} 
        c!{\vrule width \arrayrulewidth} 
        >{\columncolor{gray!25}}c!{\vrule width \arrayrulewidth} 
        c!{\vrule width \arrayrulewidth} 
        >{\columncolor{gray!25}}c!{\vrule width \arrayrulewidth} 
        c!{\vrule width \arrayrulewidth} 
        >{\columncolor{gray!25}}c}
        \toprule
        \multicolumn{2}{c}{} & \multicolumn{2}{c}{} & \multicolumn{6}{c}{\textbf{ZS-XLT}} \\ \cmidrule(lr){5-10}
        \multicolumn{1}{c}{} & \multicolumn{1}{c}{} & \multicolumn{2}{c}{\textbf{EN (1)}} & \multicolumn{2}{c}{\textbf{S4T (89)}} & \multicolumn{2}{c}{\textbf{N/A (12)}} & \multicolumn{2}{c}{\textbf{AVG (101)}} \\
        \cmidrule(lr){3-4} \cmidrule(lr){5-6} \cmidrule(lr){7-8} \cmidrule(lr){9-10}
        \textbf{\textsc{S4Tv2-L}} &  \multicolumn{1}{c!{\vrule width \arrayrulewidth}}{\textbf{Size}} & \multicolumn{1}{c!{\vrule width \arrayrulewidth}}{\textbf{B}} & \textbf{W} & \textbf{B} & \textbf{W} & \textbf{B} & \textbf{W} & \textbf{B} & \textbf{W} \\ \hline \rule{0pt}{2.5ex}
        \textsc{Incl. LA}  &  635M         & 87.4 & 88.5 & 82.2 & 79.1 & 55.7 & 52.9 & 79.0 & 75.8 \\
        \textsc{Excl. LA}  &  588M         & 88.5 & 86.4 & 80.9 & 77.5 & 55.6 & 52.0 & 77.8 & 74.4 \\ \hline \rule{0pt}{2.5ex}
        $\Delta$           &   47M         & +1.1 & -2.1 & -1.3 & -2.6 & -0.1 & -0.9 & -1.2 & -1.4 \\
        \bottomrule
    \end{tabular}
    \end{center}
\caption{\textbf{Ablation of Length Adaptor.} We benchmark SeamlessM4Tv2-Large with and without the Length Adaptor on SIB-Fleurs. See Table \ref{tab:main-results} for further details.}
\vspace{-0.5cm}
\label{tab:s4t-length-adaptor}
\end{table}

\section{Conclusion}
\label{sec:conclusion}
We introduce Fleurs-SLU, a multilingual SLU benchmark for semantic speech classification across 102 languages and multiple-choice question answering from spoken paragraphs in 92 languages. Using Fleurs-SLU, we evaluate massively multilingual speech models in both end-to-end speech classification and a cascaded approach that combines initial speech-to-text transcription and subsequent text-based classification with LLMs. Our findings indicate that, while cascaded systems remain the most robust option, multilingual speech encoders can achieve competitive performance when adequately pre-trained. Moreover, speech-LLMs can yield state-of-the-art performance on par with the best targeted SLU pipelines when appropriately aligned for multilingual spoken language instruction following. Moreover, speech-LLMs can achieve state-of-the-art performance on par or better than the best targeted SLU pipelines when properly aligned for multilingual spoken language instruction following. Furthermore, we observe a strong correlation between strong multilingual SLU and both the robustness of multilingual ASR and the effectiveness of cross-modal speech translation to English text. This suggests that multilingual SLU and multilingual ASR can be mutually beneficial. We hope that our findings inspire future work towards developing more efficient multilingual speech encoders that are jointly pre-trained for both multilingual ASR and SLU to close the performance gap between end-to-end speech classification and cascaded approaches.


\bibliography{custom,anthology}
\bibliographystyle{colm2025_conference}

\appendix
\section{Appendix}
\label{sec:appendix}
\subsection{Further Experimental Details}

\rparagraph{Compute infrastructure} All experiments run on a single Nvidia L40S 48GB or A100 80GB, respectively. We estimate that the total compute budget accumulates to about 2K GPU hours. Training runs on SIB-Fleurs require about 30-45 minutes, while fine-tuning LLMs for Belebele require roughly 9 hours per run. Evaluation, per checkpoint, across all languages supported by the corresponding benchmark, requires about 1 hour due to the comprehensive setups (e.g., original text and two types of transcriptions for speech recordings). 

\rparagraph{Prompting Qwen 2.5 7B-Omni \& Gemini 2-Flash} 
\label{appendix:speech-llms}

For the multimodal speech-LLMs Qwen 2.5 7B-Omni and Gemini 2-Flash, we apply the following utterance pre-processing. Each audio segment is normalized to a target root mean square (RMS) level of 0.07 to ensure consistent loudness across samples and reduce variability due to recording conditions. Specifically, we first normalize the loudness of individual utterances; for Belebele-Fleurs, we additionally normalize the loudness after concatenation. The models are then prompted using the task- and modality-specific templates detailed below.

\begin{figure*}[ht]

\begin{tcolorbox}[
    colback=white!95!black,
    colframe=black,
    title=SIB-Fleurs,
    fonttitle=\bfseries,
    boxrule=0.5pt,
    arc=4pt,
    outer arc=4pt,
    width=\textwidth,
    enlarge left by=0mm,
    enlarge right by=0mm,
    before skip=1em,
    after skip=1em,
]

\rparagraph{Speech} \\
The utterance belongs to one of the following topics.\\
Utterance: \texttt{[IN-LANGUAGE UTTERANCE]} \\

\rparagraph{Text} \\
The sentence belongs to one of the following topics.\\
Sentence: \texttt{[IN-LANGUAGE SENTENCE]}

\medskip

Topics:
\begin{enumerate}
    \item Entertainment
    \item Geography
    \item Health
    \item Politics
    \item Science and Technology
    \item Sports
    \item Travel
\end{enumerate}

\medskip

Please respond with only the number of the correct answer (1, 2, 3, 4, 5, 6, or 7).

\end{tcolorbox}

\begin{tcolorbox}[
    colback=white!95!black,
    colframe=black,
    title=Belebele-Fleurs,
    fonttitle=\bfseries,
    boxrule=0.5pt,
    arc=4pt,
    outer arc=4pt,
    width=\textwidth,
    enlarge left by=0mm,
    enlarge right by=0mm,
    before skip=1em,
    after skip=1em,
]

\rparagraph{Speech}\\
Listen to the audio passage. Based on the audio, answer the following multiple-choice question. \\
Passage: \texttt{[IN-LANGUAGE CONCATENATED UTTERANCES]}

\medskip

\rparagraph{Text}\\
Given the paragraph, answer the following multiple-choice question.\\
Paragraph: \texttt{[IN-LANGUAGE PARAGRAPH]}

\medskip

Question: \texttt{[IN-LANGUAGE QUESTION]}

\medskip

Options:
\begin{enumerate}
    \item \texttt{[IN-LANGUAGE MULTIPLE-CHOICE-ANSWER 1]}
    \item \texttt{[IN-LANGUAGE MULTIPLE-CHOICE-ANSWER 2]}
    \item \texttt{[IN-LANGUAGE MULTIPLE-CHOICE-ANSWER 3]}
    \item \texttt{[IN-LANGUAGE MULTIPLE-CHOICE-ANSWER 4]}
\end{enumerate}

Please respond with only the number of the correct answer (1, 2, 3, or 4).

\end{tcolorbox}
\end{figure*}


\begin{table*}[ht]
\subsection{SIB-Fleurs}
\label{subsec:sib-samples}
    \begin{minipage}{0.48\textwidth}
        \centering
        \renewcommand{\arraystretch}{0.8} 
        \rowcolors{2}{gray!15}{white} 
        \begin{adjustbox}{max height=0.6\textheight, max width=\columnwidth}
        \begin{tabular}{l l c c c}
                \textbf{Language Code} & \textbf{Language Name}  & \textbf{Train} & \textbf{Validation} & \textbf{Test} \\ \midrule \midrule
                
                \texttt{afr\_Latn} & Afrikaans & 406 & 86 & 95 \\
                \texttt{amh\_Ethi} & Amharic & 752 & 54 & 149 \\
                \texttt{arb\_Arab} & Modern Standard & 579 & 64 & 133 \\
                \texttt{asm\_Beng} & Assamese & 730 & 71 & 176 \\
                \texttt{ast\_Latn} & Asturian & 701 & 69 & 177 \\
                \texttt{azj\_Latn} & North Azerbaijani & 712 & 71 & 174 \\
                \texttt{bel\_Cyrl} & Belarusian & 690 & 71 & 177 \\
                \texttt{ben\_Beng} & Bengali & 742 & 71 & 176 \\
                \texttt{bos\_Latn} & Bosnian & 746 & 71 & 177 \\
                \texttt{bul\_Cyrl} & Bulgarian & 749 & 70 & 176 \\
                \texttt{cat\_Latn} & Catalan & 683 & 71 & 177 \\
                \texttt{ceb\_Latn} & Cebuano & 741 & 61 & 149 \\
                \texttt{ces\_Latn} & Czech & 732 & 68 & 172 \\
                \texttt{ckb\_Arab} & Central Kurdish & 738 & 70 & 176 \\
                \texttt{cym\_Latn} & Welsh & 739 & 71 & 177 \\
                \texttt{dan\_Latn} & Danish & 696 & 70 & 177 \\
                \texttt{deu\_Latn} & German & 736 & 69 & 175 \\
                \texttt{ell\_Grek} & Greek & 750 & 67 & 168 \\
                \texttt{eng\_Latn} & English & 738 & 71 & 177 \\
                \texttt{est\_Latn} & Estonian & 700 & 71 & 176 \\
                \texttt{fin\_Latn} & Finnish & 735 & 71 & 175 \\
                \texttt{fra\_Latn} & French & 753 & 65 & 164 \\
                \texttt{fuv\_Latn} & Nigerian Fulfulde & 752 & 68 & 166 \\
                \texttt{gaz\_Latn} & West Central Oromo & 574 & 6 & 17 \\
                \texttt{gle\_Latn} & Irish & 731 & 71 & 176 \\
                \texttt{glg\_Latn} & Galician & 660 & 71 & 174 \\
                \texttt{guj\_Gujr} & Gujarati & 752 & 71 & 177 \\
                \texttt{hau\_Latn} & Hausa & 753 & 70 & 166 \\
                \texttt{heb\_Hebr} & Hebrew & 754 & 70 & 175 \\
                \texttt{hin\_Deva} & Hindi & 653 & 60 & 132 \\
                \texttt{hrv\_Latn} & Croatian & 756 & 71 & 176 \\
                \texttt{hun\_Latn} & Hungarian & 750 & 71 & 177 \\
                \texttt{hye\_Armn} & Armenian & 741 & 71 & 177 \\
                \texttt{ibo\_Latn} & Igbo & 737 & 71 & 177 \\
                \texttt{ind\_Latn} & Indonesian & 728 & 69 & 167 \\
                \texttt{isl\_Latn} & Icelandic & 381 & 18 & 23 \\
                \texttt{ita\_Latn} & Italian & 743 & 69 & 175 \\
                \texttt{jav\_Latn} & Javanese & 740 & 67 & 171 \\
                \texttt{jpn\_Jpan} & Japanese & 662 & 62 & 164 \\
                \texttt{kam\_Latn} & Kamba & 752 & 69 & 179 \\
                \texttt{kan\_Knda} & Kannada & 660 & 70 & 174 \\
                \texttt{kat\_Geor} & Georgian & 557 & 69 & 177 \\
                \texttt{kaz\_Cyrl} & Kazakh & 749 & 70 & 176 \\
                \texttt{kea\_Latn} & Kabuverdianu & 725 & 71 & 175 \\
                \texttt{khk\_Cyrl} & Halh & 743 & 71 & 177 \\
                \texttt{khm\_Khmr} & Khmer & 588 & 69 & 168 \\
                \texttt{kir\_Cyrl} & Kyrgyz & 729 & 71 & 177 \\
                \texttt{kor\_Hang} & Korean & 669 & 61 & 141 \\
                \texttt{lao\_Laoo} & Lao & 591 & 54 & 132 \\
                \texttt{lin\_Latn} & Lingala & 755 & 59 & 139 \\
                \texttt{lit\_Latn} & Lithuanian & 730 & 71 & 178 \\
                                    
                 \bottomrule
            \end{tabular}
        \end{adjustbox}
    \end{minipage}
    \hfill
    \begin{minipage}{0.48\textwidth}
        \centering
        \renewcommand{\arraystretch}{0.8} 
        \rowcolors{2}{gray!15}{white} 
        \begin{adjustbox}{max height=0.6\textheight, max width=\columnwidth}
            \begin{tabular}{l l c c c}
                \textbf{Language Code} & \textbf{Language Name}  & \textbf{Train} & \textbf{Validation} & \textbf{Test} \\ \midrule
                \midrule
                \texttt{ltz\_Latn} & Luxembourgish & 703 & 71 & 176 \\
                \texttt{lug\_Latn} & Ganda & 691 & 70 & 173 \\
                \texttt{luo\_Latn} & Luo & 698 & 39 & 98 \\
                \texttt{lvs\_Latn} & Standard Latvian & 634 & 69 & 174 \\
                \texttt{mal\_Mlym} & Malayalam & 723 & 68 & 174 \\
                \texttt{mar\_Deva} & Marathi & 749 & 71 & 177 \\
                \texttt{mkd\_Cyrl} & Macedonian & 680 & 71 & 177 \\
                \texttt{mlt\_Latn} & Maltese & 731 & 71 & 176 \\
                \texttt{mri\_Latn} & Maori & 749 & 71 & 176 \\
                \texttt{mya\_Mymr} & Burmese & 746 & 71 & 175 \\
                \texttt{nld\_Latn} & Dutch & 729 & 58 & 123 \\
                \texttt{nob\_Latn} & Norwegian Bokmål & 723 & 51 & 127 \\
                \texttt{npi\_Deva} & Nepali & 754 & 70 & 175 \\
                \texttt{nso\_Latn} & Northern Sotho & 633 & 70 & 169 \\
                \texttt{nya\_Latn} & Nyanja & 720 & 68 & 169 \\
                \texttt{oci\_Latn} & Occitan & 756 & 71 & 177 \\
                \texttt{ory\_Orya} & Odia & 442 & 71 & 168 \\
                \texttt{pan\_Guru} & Eastern Panjabi & 580 & 56 & 143 \\
                \texttt{pbt\_Arab} & Southern Pashto & 701 & 55 & 144 \\
                \texttt{pes\_Arab} & Western Persian & 692 & 66 & 165 \\
                \texttt{pol\_Latn} & Polish & 723 & 68 & 165 \\
                \texttt{por\_Latn} & Portuguese & 728 & 70 & 177 \\
                \texttt{ron\_Latn} & Romanian & 734 & 69 & 177 \\
                \texttt{rus\_Cyrl} & Russian & 733 & 71 & 173 \\
                \texttt{slk\_Latn} & Slovak & 628 & 71 & 169 \\
                \texttt{slv\_Latn} & Slovenian & 704 & 71 & 174 \\
                \texttt{sna\_Latn} & Shona & 689 & 71 & 176 \\
                \texttt{snd\_Arab} & Sindhi & 749 & 71 & 177 \\
                \texttt{som\_Latn} & Somali & 746 & 70 & 177 \\
                \texttt{spa\_Latn} & Spanish & 676 & 71 & 177 \\
                \texttt{srp\_Cyrl} & Serbian & 730 & 63 & 164 \\
                \texttt{swe\_Latn} & Swedish & 686 & 71 & 168 \\
                \texttt{swh\_Latn} & Swahili & 745 & 65 & 154 \\
                \texttt{tam\_Taml} & Tamil & 693 & 71 & 169 \\
                \texttt{tel\_Telu} & Telugu & 658 & 66 & 153 \\
                \texttt{tgk\_Cyrl} & Tajik & 680 & 69 & 163 \\
                \texttt{tgl\_Latn} & Tagalog & 604 & 71 & 176 \\
                \texttt{tha\_Thai} & Thai & 710 & 71 & 176 \\
                \texttt{tur\_Latn} & Turkish & 692 & 67 & 164 \\
                \texttt{ukr\_Cyrl} & Ukrainian & 732 & 67 & 164 \\
                \texttt{umb\_Latn} & Umbundu & 473 & 39 & 108 \\
                \texttt{urd\_Arab} & Urdu & 636 & 65 & 120 \\
                \texttt{uzn\_Latn} & Northern Uzbek & 734 & 69 & 175 \\
                \texttt{vie\_Latn} & Vietnamese & 737 & 70 & 176 \\
                \texttt{wol\_Latn} & Wolof & 643 & 52 & 123 \\
                \texttt{xho\_Latn} & Xhosa & 756 & 71 & 177 \\
                \texttt{yor\_Latn} & Yoruba & 686 & 71 & 172 \\
                \texttt{zho\_Hans} & Chinese & 751 & 71 & 176 \\ 
                \texttt{zho\_Hant} & Chinese & 624 & 70 & 172 \\
                \texttt{zsm\_Latn} & Standard Malay & 713 & 67 & 171 \\
                \texttt{zul\_Latn} & Zulu & 739 & 69 & 175 \\
                \bottomrule
            \end{tabular}
        \end{adjustbox}
    \end{minipage}
    \caption{Number of samples by split and language in SIB-Fleurs.}
    \label{tab:fleurs-sib-statistics}
\end{table*}

\begin{table*}[ht]
\subsection{Belebele-Fleurs}
\label{subsec:belebele-fleurs}

    \begin{minipage}{0.48\textwidth}
        \centering
        \renewcommand{\arraystretch}{0.8} 
        \rowcolors{2}{gray!15}{white} 
        \begin{adjustbox}{max height=0.6\textheight, max width=\columnwidth}
            \begin{tabular}{l l c}
                \textbf{Language Code} & \textbf{Language Name}  & \textbf{Samples} \\ \midrule \midrule

                    \texttt{afr\_Latn} & Afrikaans  & 309 \\
                    \texttt{amh\_Ethi} & Amharic  & 782 \\
                    \texttt{arb\_Arab} & Modern Standard  & 387 \\
                    \texttt{asm\_Beng} & Assamese  & 824 \\
                    \texttt{azj\_Latn} & North Azerbaijani  & 759 \\
                    \texttt{ben\_Beng} & Bengali  & 855 \\
                    \texttt{bul\_Cyrl} & Bulgarian  & 873 \\
                    \texttt{cat\_Latn} & Catalan  & 652 \\
                    \texttt{ceb\_Latn} & Cebuano  & 783 \\
                    \texttt{ces\_Latn} & Czech  & 802 \\
                    \texttt{ckb\_Arab} & Central Kurdish  & 842 \\
                    \texttt{dan\_Latn} & Danish  & 696 \\
                    \texttt{deu\_Latn} & German  & 804 \\
                    \texttt{ell\_Grek} & Greek  & 837 \\
                    \texttt{eng\_Latn} & English  & 844 \\
                    \texttt{est\_Latn} & Estonian  & 736 \\
                    \texttt{fin\_Latn} & Finnish  & 826 \\
                    \texttt{fra\_Latn} & French  & 839 \\
                    \texttt{fuv\_Latn} & Nigerian Fulfulde  & 848 \\
                    \texttt{gaz\_Latn} & West Central Oromo  & 25  \\
                    \texttt{guj\_Gujr} & Gujarati  & 880 \\
                    \texttt{hau\_Latn} & Hausa  & 838 \\
                    \texttt{heb\_Hebr} & Hebrew  & 878 \\
                    \texttt{hin\_Deva} & Hindi  & 515 \\
                    \texttt{hrv\_Latn} & Croatian  & 896 \\
                    \texttt{hun\_Latn} & Hungarian  & 879 \\
                    \texttt{hye\_Armn} & Armenian  & 861 \\
                    \texttt{ibo\_Latn} & Igbo  & 838 \\
                    \texttt{ind\_Latn} & Indonesian  & 783 \\
                    \texttt{isl\_Latn} & Icelandic  & 81  \\
                    \texttt{ita\_Latn} & Italian  & 851 \\
                    \texttt{jav\_Latn} & Javanese  & 835 \\
                    \texttt{jpn\_Jpan} & Japanese  & 590 \\
                    \texttt{kan\_Knda} & Kannada  & 606 \\
                    \texttt{kat\_Geor} & Georgian  & 372 \\
                    \texttt{kaz\_Cyrl} & Kazakh  & 870 \\
                    \texttt{kea\_Latn} & Kabuverdianu  & 770 \\
                    \texttt{khk\_Cyrl} & Halh  & 869 \\
                    \texttt{khm\_Khmr} & Khmer  & 439 \\
                    \texttt{kir\_Cyrl} & Kyrgyz  & 811 \\
                    \texttt{kor\_Hang} & Korean  & 535 \\
                    \texttt{lao\_Laoo} & Lao  & 346 \\
                    \texttt{lin\_Latn} & Lingala  & 778 \\
                    \texttt{lit\_Latn} & Lithuanian  & 834 \\
                    \texttt{lug\_Latn} & Ganda  & 703 \\
                    \texttt{luo\_Latn} & Luo  & 512 \\
                    \texttt{lvs\_Latn} & Standard Latvian & 555 \\
                    \texttt{mal\_Mlym} & Malayalam & 809 \\
                    \bottomrule

                \end{tabular}
            \end{adjustbox}
        \end{minipage}
        \hfill
        \begin{minipage}{0.48\textwidth}
            \centering
            \renewcommand{\arraystretch}{0.8} 
            \rowcolors{2}{gray!15}{white} 
            \begin{adjustbox}{max height=0.6\textheight, max width=\columnwidth}
            \begin{tabular}{l l c}
                \textbf{Language} & \textbf{Name} & \textbf{Samples} \\ \midrule \midrule
                    \texttt{mar\_Deva} & Marathi & 869 \\
                    \texttt{mkd\_Cyrl} & Macedonian & 667 \\
                    \texttt{mlt\_Latn} & Maltese & 816 \\
                    \texttt{mri\_Latn} & Maori & 877 \\
                    \texttt{mya\_Mymr} & Burmese & 864 \\
                    \texttt{nld\_Latn} & Dutch & 674 \\
                    \texttt{nob\_Latn} & Norwegian Bokmål & 635 \\
                    \texttt{npi\_Deva} & Nepali & 876 \\
                    \texttt{nso\_Latn} & Northern Sotho & 569 \\
                    \texttt{nya\_Latn} & Nyanja & 752 \\
                    \texttt{ory\_Orya} & Odia & 220 \\
                    \texttt{pan\_Guru} & Eastern Panjabi & 396 \\
                    \texttt{pbt\_Arab} & Southern Pashto & 628 \\
                    \texttt{pes\_Arab} & Western Persian & 673 \\
                    \texttt{pol\_Latn} & Polish & 765 \\
                    \texttt{por\_Latn} & Portuguese & 791 \\
                    \texttt{ron\_Latn} & Romanian & 815 \\
                    \texttt{rus\_Cyrl} & Russian & 819 \\
                    \texttt{slk\_Latn} & Slovak & 513 \\
                    \texttt{slv\_Latn} & Slovenian & 724 \\
                    \texttt{sna\_Latn} & Shona & 735 \\
                    \texttt{snd\_Arab} & Sindhi & 878 \\
                    \texttt{som\_Latn} & Somali & 874 \\
                    \texttt{spa\_Latn} & Spanish & 659 \\
                    \texttt{srp\_Cyrl} & Serbian & 766 \\
                    \texttt{swe\_Latn} & Swedish & 681 \\
                    \texttt{swh\_Latn} & Swahili & 780 \\
                    \texttt{tam\_Taml} & Tamil & 714 \\
                    \texttt{tel\_Telu} & Telugu & 567 \\
                    \texttt{tgk\_Cyrl} & Tajik & 632 \\
                    \texttt{tgl\_Latn} & Tagalog & 505 \\
                    \texttt{tha\_Thai} & Thai & 745 \\
                    \texttt{tur\_Latn} & Turkish & 706 \\
                    \texttt{ukr\_Cyrl} & Ukrainian & 773 \\
                    \texttt{urd\_Arab} & Urdu & 482 \\
                    \texttt{uzn\_Latn} & Northern Uzbek & 812 \\
                    \texttt{vie\_Latn} & Vietnamese & 847 \\
                    \texttt{wol\_Latn} & Wolof & 495 \\
                    \texttt{xho\_Latn} & Xhosa & 900 \\
                    \texttt{yor\_Latn} & Yoruba & 652 \\
                    \texttt{zho\_Hans} & Chinese & 888 \\
                    \texttt{zho\_Hant} & Chinese & 527 \\
                    \texttt{zsm\_Latn} & Standard Malay & 749 \\
                    \texttt{zul\_Latn} & Zulu & 838 \\ \bottomrule
            \end{tabular}
        \end{adjustbox}
    \end{minipage}
    \caption{Number of samples per language in Belebele-Fleurs.}
    \label{tab:fleurs-belebele-statistics}
\end{table*}

\begin{table*}[ht]
    \subsection{Analysis of Speech-To-English-Text-Translation Performance}
    \label{subsec:s2ett-full-results}
    \begin{minipage}{0.48\textwidth}
        \centering
        \renewcommand{\arraystretch}{0.8} 
        \rowcolors{2}{gray!15}{white} 
        \begin{adjustbox}{max height=0.6\textheight, max width=\columnwidth}
            \begin{tabular}{lcc}
            \toprule
             \textbf{Language} & \textbf{Whisper-v3-Large} & \textbf{SeamlessM4Tv2-Large} \\
            \midrule
             \textbf{\texttt{AVG}} & \textbf{13.4} & \textbf{23.6} \\
                \texttt{afr\_Latn} & 32.9 & 42.5 \\
                \texttt{amh\_Ethi} & 0.8 & 18.5 \\
                \texttt{arb\_Arab} & 19.3 & 33.4 \\
                \texttt{asm\_Beng} & 2.3 & 21.9 \\
                \texttt{ast\_Latn} & 25.7 & 27.7 \\
                \texttt{azj\_Latn} & 10.7 & 18.3 \\
                \texttt{bel\_Cyrl} & 10.6 & 17.9 \\
                \texttt{ben\_Beng} & 8.1 & 27.2 \\
                \texttt{bos\_Latn} & 27.6 & 35.6 \\
                \texttt{bul\_Cyrl} & 26.4 & 33.2 \\
                \texttt{cat\_Latn} & 32.5 & 39.9 \\
                \texttt{ceb\_Latn} & 6.2 & 9.1 \\
                \texttt{ces\_Latn} & 24.5 & 33.1 \\
                \texttt{ckb\_Arab} & 1.6 & 23.2 \\
                \texttt{cym\_Latn} & 9.2 & 33.2 \\
                \texttt{dan\_Latn} & 32.6 & 38.9 \\
                \texttt{deu\_Latn} & 32.6 & 36.8 \\
                \texttt{ell\_Grek} & 20.5 & 27.7 \\
                \texttt{est\_Latn} & 15.3 & 30.4 \\
                \texttt{fin\_Latn} & 19.2 & 27.4 \\
                \texttt{fra\_Latn} & 34.0 & 36.1 \\
                \texttt{fuv\_Latn} & 0.2 & 0.9 \\
                \texttt{gaz\_Latn} & 0.3 & 0.7 \\
                \texttt{gle\_Latn} & 1.5 & 18.6 \\
                \texttt{glg\_Latn} & 27.9 & 35.6 \\
                \texttt{guj\_Gujr} & 10.7 & 30.2 \\
                \texttt{hau\_Latn} & 0.5 & 1.3 \\
                \texttt{heb\_Hebr} & 16.1 & 31.3 \\
                \texttt{hin\_Deva} & 19.3 & 28.5 \\
                \texttt{hrv\_Latn} & 24.9 & 31.7 \\
                \texttt{hun\_Latn} & 17.5 & 27.9 \\
                \texttt{hye\_Armn} & 9.0 & 31.2 \\
                \texttt{ibo\_Latn} & 0.6 & 2.6 \\
                \texttt{ind\_Latn} & 26.3 & 30.7 \\
                \texttt{isl\_Latn} & 7.5 & 25.7 \\
                \texttt{ita\_Latn} & 23.8 & 27.6 \\
                \texttt{jav\_Latn} & 4.1 & 23.6 \\
                \texttt{jpn\_Jpan} & 16.0 & 17.8 \\
                \texttt{kam\_Latn} & 0.8 & 2.5 \\
                \texttt{kan\_Knda} & 6.7 & 24.9 \\
                \texttt{kat\_Geor} & 2.3 & 21.7 \\
                \texttt{kaz\_Cyrl} & 3.8 & 25.0 \\
                \texttt{kea\_Latn} & 26.5 & 28.7 \\
                \texttt{khk\_Cyrl} & 0.8 & 18.5 \\
                \texttt{khm\_Khmr} & 4.5 & 22.3 \\
                \texttt{kir\_Cyrl} & 2.4 & 19.0 \\
                \texttt{kor\_Hang} & 19.0 & 22.7 \\
                \texttt{lao\_Laoo} & 7.1 & 25.9 \\
                \texttt{lin\_Latn} & 0.5 & 1.5 \\
                \texttt{lit\_Latn} & 12.7 & 24.3 \\
            \bottomrule
            \end{tabular}
        \end{adjustbox}
    \end{minipage}
    \hfill
    \begin{minipage}{0.48\textwidth}
        \centering
        \renewcommand{\arraystretch}{0.8} 
        \rowcolors{2}{gray!15}{white} 
        \begin{adjustbox}{max height=0.6\textheight, max width=\columnwidth}
            \begin{tabular}{lcc}
                \toprule
                \textbf{Language} & \textbf{Whisper-v3-Large} & \textbf{SeamlessM4Tv2-Large} \\
                \midrule
                \texttt{ltz\_Latn} & 16.0 & 17.6 \\
                \texttt{lug\_Latn} & 0.7 & 18.3 \\
                \texttt{luo\_Latn} & 0.9 & 1.3 \\
                \texttt{lvs\_Latn} & 13.3 & 29.0 \\
                \texttt{mal\_Mlym} & 9.9 & 25.0 \\
                \texttt{mar\_Deva} & 9.9 & 27.2 \\
                \texttt{mkd\_Cyrl} & 26.0 & 35.6 \\
                \texttt{mlt\_Latn} & 11.1 & 40.1 \\
                \texttt{mri\_Latn} & 6.5 & 1.3 \\
                \texttt{mya\_Mymr} & 0.4 & 19.4 \\
                \texttt{nld\_Latn} & 22.1 & 25.8 \\
                \texttt{nob\_Latn} & 29.7 & 34.7 \\
                \texttt{npi\_Deva} & 10.6 & 9.4 \\
                \texttt{nso\_Latn} & 0.7 & 2.6 \\
                \texttt{nya\_Latn} & 0.9 & 19.3 \\
                \texttt{oci\_Latn} & 17.6 & 23.5 \\
                \texttt{ory\_Orya} & 4.8 & 26.8 \\
                \texttt{pan\_Guru} & 12.9 & 30.2 \\
                \texttt{pbt\_Arab} & 1.7 & 18.2 \\
                \texttt{pes\_Arab} & 16.0 & 30.5 \\
                \texttt{pol\_Latn} & 20.7 & 24.2 \\
                \texttt{por\_Latn} & 37.5 & 38.9 \\
                \texttt{ron\_Latn} & 30.6 & 36.5 \\
                \texttt{rus\_Cyrl} & 26.2 & 30.3 \\
                \texttt{slk\_Latn} & 25.1 & 33.6 \\
                \texttt{slv\_Latn} & 18.0 & 27.4 \\
                \texttt{sna\_Latn} & 1.0 & 3.8 \\
                \texttt{snd\_Arab} & 3.6 & 8.7 \\
                \texttt{som\_Latn} & 0.4 & 18.1 \\
                \texttt{spa\_Latn} & 22.0 & 24.6 \\
                \texttt{srp\_Cyrl} & 29.8 & 37.3 \\
                \texttt{swe\_Latn} & 34.3 & 38.5 \\
                \texttt{swh\_Latn} & 5.6 & 32.4 \\
                \texttt{tam\_Taml} & 6.0 & 22.5 \\
                \texttt{tel\_Telu} & 10.3 & 26.5 \\
                \texttt{tgk\_Cyrl} & 8.7 & 26.9 \\
                \texttt{tgl\_Latn} & 21.2 & 25.6 \\
                \texttt{tha\_Thai} & 13.4 & 23.5 \\
                \texttt{tur\_Latn} & 22.2 & 30.1 \\
                \texttt{ukr\_Cyrl} & 27.6 & 32.8 \\
                \texttt{umb\_Latn} & 0.2 & 1.1 \\
                \texttt{urd\_Arab} & 15.9 & 25.5 \\
                \texttt{uzn\_Latn} & 5.0 & 25.5 \\
                \texttt{vie\_Latn} & 19.2 & 26.4 \\
                \texttt{wol\_Latn} & 1.2 & 1.6 \\
                \texttt{xho\_Latn} & 0.8 & 7.6 \\
                \texttt{yor\_Latn} & 0.6 & 14.6 \\
                \texttt{zho\_Hans} & 14.6 & 22.4 \\
                \texttt{zho\_Hant} & 8.7 & 18.5 \\
                \texttt{zsm\_Latn} & 24.7 & 31.0 \\
                \texttt{zul\_Latn} & 0.6 & 12.4 \\
                \bottomrule
            \end{tabular}
        \end{adjustbox}
    \end{minipage}
\caption{Per-language average sacreBLEU for Speech-to-English-Text-Translation of Fleurs utterances to their original English sentences, computed over the pooled train, dev, and test splits. Results are shown for Whisper-v3-Large and SeamlessM4Tv2-Large (cf. \S\ref{sec:experimental-setup}).}
\end{table*}

\begin{table*}[ht]
    \subsection{Analysis of ASR Performance}
    \label{subsec:asr-full-results}
    \begin{minipage}{0.48\textwidth}
        \centering
        \renewcommand{\arraystretch}{0.8} 
        \rowcolors{2}{gray!15}{white} 
        \begin{adjustbox}{max height=0.6\textheight, max width=\columnwidth}
            \begin{tabular}{lcc}
            \toprule
                \textbf{Language} & \textbf{Whisper-v3-Large} & \textbf{SeamlessM4Tv2-Large} \\
                \midrule
                \texttt{afr\_Latn} & 13.1$_\text{8.8}$ & 12.6$_\text{13.6}$ \\
                \texttt{amh\_Ethi} & 207.8$_\text{97.7}$ & 23.9$_\text{13.0}$ \\
                \texttt{arb\_Arab} & 7.6$_\text{7.8}$ & 8.0$_\text{11.3}$ \\
                \texttt{asm\_Beng} & 99.0$_\text{34.9}$ & 15.1$_\text{9.2}$ \\
                \texttt{ast\_Latn} & 15.5$_\text{6.8}$ & 19.8$_\text{15.6}$ \\
                \texttt{azj\_Latn} & 7.2$_\text{9.4}$ & 8.3$_\text{9.6}$ \\
                \texttt{bel\_Cyrl} & 11.9$_\text{5.6}$ & 6.9$_\text{10.8}$ \\
                \texttt{ben\_Beng} & 33.9$_\text{34.9}$ & 9.0$_\text{6.4}$ \\
                \texttt{bos\_Latn} & 4.6$_\text{5.2}$ & 7.8$_\text{11.4}$ \\
                \texttt{bul\_Cyrl} & 5.2$_\text{5.9}$ & 9.6$_\text{13.7}$ \\
                \texttt{cat\_Latn} & 3.3$_\text{4.6}$ & 5.3$_\text{9.4}$ \\
                \texttt{ceb\_Latn} & 17.4$_\text{47.7}$ & 19.0$_\text{11.5}$ \\
                \texttt{ces\_Latn} & 4.2$_\text{10.0}$ & 8.0$_\text{11.8}$ \\
                \texttt{ckb\_Arab} & 98.3$_\text{140.7}$ & 13.9$_\text{15.7}$ \\
                \texttt{cym\_Latn} & 15.8$_\text{18.6}$ & 14.4$_\text{19.6}$ \\
                \texttt{dan\_Latn} & 4.8$_\text{5.2}$ & 10.0$_\text{13.2}$ \\
                \texttt{deu\_Latn} & 2.2$_\text{3.3}$ & 6.5$_\text{9.7}$ \\
                \texttt{ell\_Grek} & 6.4$_\text{7.4}$ & 10.8$_\text{12.5}$ \\
                \texttt{eng\_Latn} & 3.1$_\text{4.9}$ & 6.2$_\text{9.1}$ \\
                \texttt{est\_Latn} & 7.3$_\text{8.7}$ & 11.0$_\text{15.6}$ \\
                \texttt{fin\_Latn} & 3.2$_\text{4.3}$ & 10.2$_\text{13.3}$ \\
                \texttt{fra\_Latn} & 2.8$_\text{4.1}$ & 6.6$_\text{9.9}$ \\
                \texttt{fuv\_Latn} & 194.3$_\text{368.0}$ & 57.0$_\text{88.7}$ \\
                \texttt{gaz\_Latn} & 35.8$_\text{14.7}$ & 47.8$_\text{40.9}$ \\
                \texttt{gle\_Latn} & 85.1$_\text{111.7}$ & 21.4$_\text{15.3}$ \\
                \texttt{glg\_Latn} & 4.3$_\text{4.0}$ & 6.5$_\text{10.3}$ \\
                \texttt{guj\_Gujr} & 21.3$_\text{17.1}$ & 10.3$_\text{8.4}$ \\
                \texttt{hau\_Latn} & 33.8$_\text{39.0}$ & 51.8$_\text{69.7}$ \\
                \texttt{heb\_Hebr} & 12.2$_\text{14.9}$ & 14.5$_\text{18.3}$ \\
                \texttt{hin\_Deva} & 11.7$_\text{15.3}$ & 10.2$_\text{10.3}$ \\
                \texttt{hrv\_Latn} & 4.5$_\text{17.2}$ & 8.5$_\text{12.5}$ \\
                \texttt{hun\_Latn} & 5.2$_\text{12.8}$ & 8.5$_\text{11.3}$ \\
                \texttt{hye\_Armn} & 18.0$_\text{32.4}$ & 9.1$_\text{10.5}$ \\
                \texttt{ibo\_Latn} & 42.9$_\text{36.4}$ & 59.3$_\text{52.3}$ \\
                \texttt{ind\_Latn} & 4.0$_\text{9.8}$ & 9.2$_\text{13.8}$ \\
                \texttt{isl\_Latn} & 16.0$_\text{26.3}$ & 10.3$_\text{12.9}$ \\
                \texttt{ita\_Latn} & 2.1$_\text{2.9}$ & 5.2$_\text{9.3}$ \\
                \texttt{jav\_Latn} & 26.3$_\text{52.3}$ & 11.3$_\text{11.0}$ \\
                \texttt{jpn\_Jpan} & 6.3$_\text{12.1}$ & 16.4$_\text{10.5}$ \\
                \texttt{kam\_Latn} & 45.1$_\text{84.7}$ & 53.9$_\text{43.7}$ \\
                \texttt{kan\_Knda} & 20.1$_\text{26.5}$ & 10.9$_\text{9.2}$ \\
                \texttt{kat\_Geor} & 19.8$_\text{13.9}$ & 7.1$_\text{9.7}$ \\
                \texttt{kaz\_Cyrl} & 8.9$_\text{8.5}$ & 9.7$_\text{12.1}$ \\
                \texttt{kea\_Latn} & 35.8$_\text{37.2}$ & 38.1$_\text{13.9}$ \\
                \texttt{khk\_Cyrl} & 37.6$_\text{36.8}$ & 14.6$_\text{22.9}$ \\
                \texttt{khm\_Khmr} & 144.0$_\text{61.7}$ & 29.8$_\text{11.2}$ \\
                \texttt{kir\_Cyrl} & 28.3$_\text{21.5}$ & 8.8$_\text{11.5}$ \\
                \texttt{kor\_Hang} & 7.4$_\text{9.0}$ & 10.3$_\text{11.3}$ \\
                \texttt{lao\_Laoo} & 109.7$_\text{45.3}$ & 30.3$_\text{11.7}$ \\
                \texttt{lin\_Latn} & 22.0$_\text{22.2}$ & 58.7$_\text{55.7}$ \\
                \texttt{lit\_Latn} & 8.3$_\text{10.6}$ & 11.9$_\text{17.8}$ \\
                \bottomrule
            \end{tabular}
        \end{adjustbox}
    \end{minipage}
    \hfill
    \begin{minipage}{0.48\textwidth}
        \renewcommand{\arraystretch}{0.8} 
        \rowcolors{2}{gray!15}{white} 
        \begin{adjustbox}{max height=0.6\textheight, max width=\columnwidth}
        \begin{tabular}{lcc}
        \toprule
            \textbf{Language} & \textbf{Whisper-v3-Large} & \textbf{SeamlessM4Tv2-Large} \\ \midrule
            \texttt{ltz\_Latn} & 29.5$_\text{16.0}$ & 40.1$_\text{15.0}$ \\
            \texttt{lug\_Latn} & 44.9$_\text{88.9}$ & 14.1$_\text{13.9}$ \\
            \texttt{luo\_Latn} & 42.7$_\text{110.1}$ & 58.8$_\text{59.5}$ \\
            \texttt{lvs\_Latn} & 7.3$_\text{20.8}$ & 8.2$_\text{13.5}$ \\
            \texttt{mal\_Mlym} & 104.3$_\text{71.4}$ & 14.2$_\text{14.1}$ \\
            \texttt{mar\_Deva} & 24.3$_\text{14.3}$ & 10.7$_\text{9.5}$ \\
            \texttt{mkd\_Cyrl} & 5.8$_\text{8.2}$ & 8.6$_\text{12.3}$ \\
            \texttt{mlt\_Latn} & 26.0$_\text{29.9}$ & 12.1$_\text{11.3}$ \\
            \texttt{mri\_Latn} & 13.4$_\text{14.7}$ & 54.7$_\text{54.4}$ \\
            \texttt{mya\_Mymr} & 132.9$_\text{78.6}$ & 22.2$_\text{9.5}$ \\
            \texttt{nld\_Latn} & 3.1$_\text{3.8}$ & 6.8$_\text{9.7}$ \\
            \texttt{nob\_Latn} & 4.6$_\text{5.7}$ & 9.0$_\text{10.2}$ \\
            \texttt{npi\_Deva} & 25.6$_\text{14.0}$ & 74.1$_\text{52.2}$ \\
            \texttt{nso\_Latn} & 95.7$_\text{162.5}$ & 66.9$_\text{63.9}$ \\
            \texttt{nya\_Latn} & 35.6$_\text{56.8}$ & 12.4$_\text{11.9}$ \\
            \texttt{oci\_Latn} & 25.9$_\text{13.9}$ & 33.8$_\text{16.5}$ \\
            \texttt{ory\_Orya} & 93.8$_\text{12.5}$ & 11.4$_\text{7.7}$ \\
            \texttt{pan\_Guru} & 44.2$_\text{49.1}$ & 10.3$_\text{8.4}$ \\
            \texttt{pbt\_Arab} & 36.8$_\text{13.9}$ & 21.6$_\text{25.3}$ \\
            \texttt{pes\_Arab} & 8.6$_\text{7.2}$ & 7.7$_\text{9.8}$ \\
            \texttt{pol\_Latn} & 2.3$_\text{3.8}$ & 7.9$_\text{13.8}$ \\
            \texttt{por\_Latn} & 2.8$_\text{4.5}$ & 7.6$_\text{10.4}$ \\
            \texttt{ron\_Latn} & 3.2$_\text{4.1}$ & 6.6$_\text{11.5}$ \\
            \texttt{rus\_Cyrl} & 2.6$_\text{4.4}$ & 6.1$_\text{10.7}$ \\
            \texttt{slk\_Latn} & 3.9$_\text{5.4}$ & 7.0$_\text{11.8}$ \\
            \texttt{slv\_Latn} & 6.0$_\text{7.4}$ & 9.2$_\text{12.2}$ \\
            \texttt{sna\_Latn} & 26.7$_\text{24.2}$ & 32.7$_\text{20.7}$ \\
            \texttt{snd\_Arab} & 94.5$_\text{36.4}$ & 33.5$_\text{17.8}$ \\
            \texttt{som\_Latn} & 34.9$_\text{28.7}$ & 17.9$_\text{18.8}$ \\
            \texttt{spa\_Latn} & 2.2$_\text{3.6}$ & 6.3$_\text{10.2}$ \\
            \texttt{srp\_Cyrl} & 76.6$_\text{26.5}$ & 7.6$_\text{12.0}$ \\
            \texttt{swe\_Latn} & 3.7$_\text{5.7}$ & 9.2$_\text{11.3}$ \\
            \texttt{swh\_Latn} & 11.0$_\text{22.2}$ & 8.9$_\text{12.0}$ \\
            \texttt{tam\_Taml} & 13.3$_\text{15.2}$ & 11.6$_\text{13.2}$ \\
            \texttt{tel\_Telu} & 90.1$_\text{91.9}$ & 13.3$_\text{14.3}$ \\
            \texttt{tgk\_Cyrl} & 28.8$_\text{37.2}$ & 8.8$_\text{9.9}$ \\
            \texttt{tgl\_Latn} & 5.0$_\text{10.3}$ & 12.0$_\text{10.8}$ \\
            \texttt{tha\_Thai} & 9.8$_\text{11.6}$ & 11.4$_\text{12.5}$ \\
            \texttt{tur\_Latn} & 6.2$_\text{25.6}$ & 7.5$_\text{9.0}$ \\
            \texttt{ukr\_Cyrl} & 3.0$_\text{4.1}$ & 8.9$_\text{12.0}$ \\
            \texttt{umb\_Latn} & 156.7$_\text{303.2}$ & 55.5$_\text{71.9}$ \\
            \texttt{urd\_Arab} & 9.4$_\text{6.9}$ & 9.6$_\text{8.0}$ \\
            \texttt{uzn\_Latn} & 25.0$_\text{32.4}$ & 8.4$_\text{11.4}$ \\
            \texttt{vie\_Latn} & 4.7$_\text{5.6}$ & 6.7$_\text{8.3}$ \\
            \texttt{wol\_Latn} & 140.6$_\text{260.8}$ & 47.9$_\text{52.8}$ \\
            \texttt{xho\_Latn} & 48.1$_\text{90.1}$ & 32.3$_\text{25.2}$ \\
            \texttt{yor\_Latn} & 49.3$_\text{18.2}$ & 33.2$_\text{13.2}$ \\
            \texttt{zho\_Hans} & 16.2$_\text{14.8}$ & 19.0$_\text{13.4}$ \\
            \texttt{zho\_Hant} & 30.6$_\text{41.5}$ & 35.1$_\text{16.8}$ \\
            \texttt{zsm\_Latn} & 3.4$_\text{4.0}$ & 8.5$_\text{10.7}$ \\
            \texttt{zul\_Latn} & 42.6$_\text{76.4}$ & 18.4$_\text{15.3}$ \\
            \bottomrule
            \end{tabular}
        \end{adjustbox}
    \end{minipage}
    \caption{Per-language average CER for transcribing Fleurs utterances to their original sentences, computed over the pooled train, dev, and test splits. Results are shown for Whisper-v3-Large and SeamlessM4Tv2-Large (cf. \S\ref{sec:experimental-setup}). Utterances in unsupported languages are transcribed using the closest manually selected supported language.}

\end{table*}

\begin{table}
    \subsection{Silent Fleurs Examples}
    \label{subsec:silent-fleurs-examples}
    \begin{centering}
        
    \renewcommand{\arraystretch}{0.8} 
    \begin{adjustbox}{max height=0.3\textheight, max width=\columnwidth}
    \rowcolors{2}{gray!15}{white} 
    \begin{tabular}{@{}lcr@{}}
        \textbf{Language} & \textbf{Split} & \textbf{Count} \\ \midrule \midrule
        \texttt{nb\_no}  & train & 497 \\
        \texttt{es\_419} & train & 490 \\
        \texttt{cy\_gb}  & train & 394 \\
        \texttt{sd\_in}  & train & 307 \\
        \texttt{ny\_mw}  & train & 15 \\
        \texttt{ckb\_iq} & train & 8 \\
        \texttt{ny\_mw}  & test  & 8 \\
        \texttt{wo\_sn}  & train & 7 \\
        \texttt{nso\_za} & test  & 6 \\
        \texttt{ny\_mw}  & dev   & 6 \\
        \texttt{ur\_pk}  & test  & 6 \\
        \texttt{ps\_af}  & train & 4 \\
        \texttt{fa\_ir}  & train & 4 \\
        \texttt{so\_so}  & train & 4 \\
        \texttt{ceb\_ph} & train & 3 \\
        \texttt{lg\_ug}  & train & 3 \\
        \texttt{kea\_cv} & train & 2 \\
        \texttt{bg\_bg}  & train & 2 \\
        \texttt{bn\_in}  & train & 2 \\
        \texttt{cy\_gb}  & test  & 2 \\
        \texttt{ff\_sn}  & train & 2 \\
        \texttt{hr\_hr}  & train & 2 \\
        \texttt{hy\_am}  & train & 2 \\
        \texttt{nso\_za} & dev   & 2 \\
        \texttt{ur\_pk}  & dev   & 2 \\
        \texttt{ar\_eg}  & train & 1 \\
        \texttt{da\_dk}  & test  & 1 \\
        \texttt{da\_dk}  & train & 1 \\
        \texttt{en\_us}  & train & 1 \\
        \texttt{ff\_sn}  & dev   & 1 \\
        \texttt{ha\_ng}  & train & 1 \\
        \texttt{he\_il}  & dev   & 1 \\
        \texttt{he\_il}  & test  & 1 \\
        \texttt{he\_il}  & train & 1 \\
        \texttt{ig\_ng}  & train & 1 \\
        \texttt{kam\_ke} & train & 1 \\
        \texttt{kn\_in}  & dev   & 1 \\
        \texttt{kn\_in}  & test  & 1 \\
        \texttt{kn\_in}  & train & 1 \\
        \texttt{mi\_nz}  & dev   & 1 \\
        \texttt{mn\_mn}  & train & 1 \\
        \texttt{ms\_my}  & train & 1 \\
        \texttt{or\_in}  & train & 1 \\
        \texttt{sk\_sk}  & dev   & 1 \\
        \texttt{sk\_sk}  & test  & 1 \\
        \texttt{so\_so}  & test  & 1 \\
        \texttt{ta\_in}  & train & 1 \\
        \texttt{te\_in}  & test  & 1 \\
        \texttt{te\_in}  & train & 1 \\
        \texttt{umb\_ao} & train & 1 \\
    \bottomrule
    \end{tabular}
    \end{adjustbox}
    \caption{Number of silent examples in Fleurs by language and split.}
    \end{centering}
    \label{tab:fleurs-silent-examples}
\end{table}

\end{document}